\documentclass[10pt,twocolumn,letterpaper]{article}

\usepackage[pagenumbers]{iccv} 

\usepackage[T1]{fontenc}

\definecolor{iccvblue}{rgb}{0.21,0.49,0.74}
\usepackage[pagebackref,breaklinks,colorlinks,allcolors=iccvblue]{hyperref}

\usepackage{threeparttable}
\usepackage{multirow} 
\usepackage{longtable}
\usepackage{diagbox}
\usepackage{multicol}
\usepackage{makecell}
\usepackage{amssymb}
\usepackage{algorithm, algorithmic}
\usepackage{setspace}
\usepackage{graphicx}
\usepackage{bbding}
\usepackage{subcaption}

\def\paperID{4502} 
\def\confName{ICCV}
\def\confYear{2025}

\title{MIBench: A Comprehensive Framework for Benchmarking Model Inversion Attack and Defense}

\renewcommand\footnotemark{}
\author{
Yixiang Qiu$^{2}$, Hongyao Yu$^{1}$, Hao Fang$^{2}$, Tianqu Zhuang$^{2}$, Wenbo Yu$^{1}$, Bin Chen$^{1\ddagger}$, \\ Xuan Wang$^{1}$, \textbf{Shu-Tao Xia$^{2}$, Ke Xu$^{3}$} \\
$^{1}$School of Computer Science and Technology, Harbin Institute of Technology, Shenzhen \\
$^{2}$Tsinghua Shenzhen International Graduate School, Tsinghua University   \\
$^{3}$Department of Computer Science and Technology, Tsinghua University \\
\footnotesize{\texttt{\{yuhongyao, yuwenbo\}@stu.hit.edu.cn;}}
\footnotesize{\texttt{\{qiu-yx24, fang-h23,zhuangtq23\}@mails.tsinghua.edu.cn;}} \\
\footnotesize{\texttt{chenbin2021@hit.edu.cn;}}
\footnotesize{\texttt{wangxuan@cs.hitsz.edu.cn;}}
\footnotesize{\texttt{xiast@sz.tsinghua.edu.cn;}}
\footnotesize{\texttt{xuke@tsinghua.edu.cn}}
\thanks{$^{\ddagger}$Corresponding author: Bin Chen (chenbin2021@hit.edu.cn).}
}

\begin{document}
\maketitle
\begingroup
\begin{abstract}
  Model Inversion (MI) attacks aim at leveraging the output information of target models to reconstruct privacy-sensitive training data, raising critical concerns regarding the privacy vulnerabilities of Deep Neural Networks (DNNs). Unfortunately, in tandem with the rapid evolution of MI attacks, the absence of a comprehensive benchmark with standardized metrics and reproducible implementations has emerged as a formidable challenge. This deficiency has hindered objective comparison of methodological advancements and reliable assessment of defense efficacy. To address this critical gap, we build the first practical benchmark named \textit{MIBench} for systematic evaluation of model inversion attacks and defenses. This benchmark bases on an extensible and reproducible modular-based toolbox which currently integrates a total of 19 state-of-the-art attack and defense methods and encompasses 9 standardized evaluation protocols. Capitalizing on this foundation, we conduct extensive evaluation from multiple perspectives to holistically compare and analyze various methods across different scenarios, such as the impact of target resolution, model predictive power, defense performance and adversarial robustness.
\end{abstract}    
\section{Introduction}

Model Inversion (MI) attacks aim to infer private training data from the output information of target models. In recent years, the proliferation of MI attacks have raised alarms over the potential privacy breaches of sensitive personal information, including the leakage of privacy images in face recognition models \citep{resnet}, sensitive health details in medical data \citep{wang2022wearable}, financial information such as transaction records and account balances \citep{ozbayoglu2020deep}, and personal preferences and social connections in social media data \citep{feng2022h}. Since \citep{zhang2020secret} first introduced the Generative Adversarial Networks (GANs) as stronger image priors and established the first GAN-based MI framework, a series of subsequent studies \citep{chen2021knowledge,yuan2023pseudo,wang2021variational,ifgmi} have driven substantial progress in this emerging field during the past few years.

Despite the remarkable development in the MI field, some critical challenges are posed due to the absence of a comprehensive and reliable benchmark. \textbf{1) Limited Reproducibility}: While most MI methods provide open-source implementations, significant reproducibility challenges persist. First, heterogeneous runtime environments \citep{ppa,plg,vmi} impede simultaneous replication across different studies. Second, the unavailability of pre-trained models in some works \citep{gmi,ked,vmi} hinders comprehensive comparative experimentation. \textbf{2) Inadequate and Unfair comparison on different methods}: The evaluation of new methods is often confined to comparisons with a narrow selection of prior works, limiting the scope and depth of analysis. Some methods \citep{lomma,plg} exhibit superior performance for lower-resolution images while other methods \citep{ppa,ifgmi} perform better at higher resolutions. However, these studies only perform evaluation under their predominant resolutions, and thus do not provide a unified and comprehensive comparison. 
These shortcomings impede both the accurate measurement of advancements in the MI field and the systematic exploration of its theoretical underpinnings, underscoring the urgent need for a harmonized framework to facilitate robust and transparent research practices.

To alleviate these problems, we establish the first comprehensive benchmark named \textit{MIBench} for MI attacks and defenses. Our proposed \textit{MIBench} is built upon an extensible modular-based toolbox, which has encompassed a total of 19 distinct attack and defense methods and coupled with 9 prevalent evaluation protocols to adequately measure the comprehensive performance of individual MI methods. Furthermore, we conduct extensive evaluation from multiple perspectives to achieve a thorough appraisal for existing MI methods, while simultaneously venturing into undiscovered insights to inspire potential avenues for future research. We expect that this reproducible and reliable benchmark will facilitate the further development of MI field and bring more innovative explorations in the subsequent study. Our main contributions are as follows:

\begin{itemize}
    \item We build the first comprehensive benchmark in MI field, featuring a reliable and reproducible modular-based toolbox and comprehensive evaluation on multiple scenarios. 
    \item We currently implement a total of 19 state-of-the-art attack methods and defense strategies and 9 standarized evaluation protocols in our toolbox.
    \item We conduct extensive experiments to thoroughly assess different MI attacks and defenses under multiple settings and investigate the effects of different factors to offer new insights on the MI field.
\end{itemize}
\section{Related Work}

\textbf{Model Inversion Attacks.} In MI attacks, the malicious adversary aims to reconstruct privacy-sensitive data by leveraging the output prediction confidence of target classifiers and other necessary auxiliary priors. Normally, the attacker requires a public dataset that shares structural similarities but has no overlapping labels with the private dataset, which is utilized to pre-train a generator before launching an attack. For example, an open-source face dataset serves as essential public data when targeting a face recognition model. In typical MI attacks based on Generative Adversarial Networks (GANs), attackers attempt to recover private images $\mathbf{x}^*$ from the GAN's latent vectors $\mathbf{z}$ initialized by Gaussian distribution, given the target image classifier $f_{\theta}$ parameterized with weights $\theta$ and the trained generator $G$. The attack process is formulated as follows:
\begin{equation}
\label{eq:basic}
    \mathbf{z}^{*} = \mathop{\arg\min}\limits_{\mathbf{z}} \mathcal{L}_{id}(f_\theta(G(\mathbf{z})),c) + \lambda\mathcal{L}_{aux}(\mathbf{z};G),
\end{equation}
where $c$ is the target class, $\mathcal{L}_{id}(\cdot,\cdot)$ typically denotes the classification loss, $\lambda$ is a hyperparameter, and $\mathcal{L}_{aux}(\cdot)$ is the prior knowledge regularization (e.g., the discriminator's classification loss) used to improve the image fidelity of $G(\mathbf{z})$. Following the above optimization, the adversary updates the latent vectors $\mathbf{z}$ into the optimal results $\mathbf{\hat{z}}$ and generate final images through $\mathbf{\hat{x}}=G(\mathbf{z^*})$.

Based on the accessibility of target models, MI attacks can be further split into \textit{white-box} attacks, \textit{black-box} attacks, and \textit{label-only} attacks. In \textit{white-box} settings, the full parameters of target models are available to attackers. GMI \citep{gmi} first proposes to incorporate the rich prior knowledge \citep{fang2023gifd, gu2020image, fang2024privacy} within the pre-trained GANs \citep{goodfellow2014generative}. Specifically, GMI starts by generating a series of preliminary fake images, and then iteratively optimizes the input latent vectors that are used for generation. Based on GMI, KEDMI \citep{ked} refines the discriminator by introducing target labels to recover the distribution of the input latent vectors. VMI \citep{vmi} utilizes variational inference to model MI attacks and adopts KL-divergence as the regularization to better estimate the target distribution. PPA \citep{ppa} introduces a series of techniques such as initial selection, post-selection, and data argumentation to enhance MI attacks and manages to recover high-resolution images by the pre-trained StyleGAN2 \citep{karras2019style}. LOMMA \citep{lomma} integrates model augmentation and model distillation into MI attacks to tackle the problem of over-fitting. PLGMI \citep{plg} leverages a top-$n$ selection technique to generate pseudo labels to further guide the training process of GAN. The most recent work IFGMI\citep{ifgmi} delves into the internal architecture of GANs and achieve higher-quality image inversion through optimization of the intermediate-layer feature vectors.

In \textit{black-box} settings, the attackers only have access to the output prediction confidence vectors (also named \textit{soft-label}). Thus, gradients of the target model can no longer be computed by the back-propagation process. 
Mirror \citep{mirror} addresses this problem by employing a genetic algorithm as an alternative to gradient descent for optimization. Following \citep{mirror}, C2FMI \citep{c2f} designs an inverse net to map soft labels into the latent space of pre-trained GANs, which serves as coarse optimization before launching the genetic algorithm. RLBMI \citep{rlb} presents a novel paradigm by incorporating reinforcement learning as the optimization mechanism.

The \textit{label-only} settings step further on the \textit{black-box} settings, relaxing the accessibility to merely one-hot vectors containing the final predicted class (also named \textit{hard-label}). BREPMI \citep{brep} estimates the optimization direction via the boundary repulsion technique. LOKT\cite{lokt} proposes to conduct MI attacks on various surrogate models instead of the unfamiliar victim model, transforming the \textit{label-only} settings into \textit{white-box} settings.

\textbf{Model Inversion Defenses.} To defend MI attacks, most existing methods can be categorized into two types according to their functional stage: \textit{train-time defense} \citep{gong2023gan, titcombe2021practical, li2022ressfl, mid, bido, ls} and \textit{inference-time defense} \citep{yang2020defending, wen2021defending, ye2022one}. \textit{Train-time defense} refers to incorporating the defense strategies during the training process. MID \cite{mid} proposes penalizing the mutual information between model inputs and outputs in the training loss to reduce the redundant information carried in the model output that may be abused by the attackers. However, simply decreasing the dependency between the inputs and outputs also results in model performance degradation. To strike a better balance between model utility and user privacy, BiDO \citep{bido} minimizes the dependency between the inputs and outputs while maximizing the dependency between the latent representations and outputs. \cite{gong2023gan} propose to leverage GAN to generate fake public samples to mislead the attackers. \cite{titcombe2021practical} defend MI attacks by adding Laplacian noise to intermediate representations. LS \citep{ls} finds that label smoothing with negative factors can help privacy preservation. TL \citep{tl} leverages transfer learning to limit the number of layers encoding sensitive information and thus improves the robustness to MI attacks. The most recent work \cite{TTS} points out that the skip connection structure of DNNs enhances MI attacks and first proposes defenses from the perspective of robust model architecture. 

\textit{Inference-time defense} refers to reducing the private information carried in the victim model's output to promote privacy without modifying the training process. This type of defense often functions as an additional plugin. Purifier \cite{yang2023purifier} proposed to train an autoencoder to purify the output vector by increasing its degree of dispersion. SSD \cite{zhuangstealthy} proposed to minimize the conditional mutual information under the constraint of bounded distortion.

\begin{table*}[!ht]
    \setlength{\tabcolsep}{5pt}
    \normalsize
    \centering
    \caption{Summary of implemented MI attack methods in our benchmark.}
    \label{table:attacks}
    \begin{threeparttable} 
    \resizebox{\linewidth}{!}{
    \begin{tabular}{ccccc}
        \toprule
         \textbf{Attack Method} & \textbf{Accessibility} & \textbf{Reference} & \textbf{GAN Prior} & \textbf{Official Resolution}\\ \midrule
         GMI \citep{gmi} & \textit{White-box} & CVPR-2020 & WGAN & $64\times64$ \\
         KEDMI \citep{ked} & \textit{White-box} & ICCV-2021 & Inversion-specific GAN$^{\dag}$ & $64\times64$ \\
         VMI \citep{vmi} & \textit{White-box} & NeurIPS-2021 & StyleGAN2 & $64\times64$ \\
         PPA \citep{ppa} & \textit{White-box} & ICML-2022 & StyleGAN2& $224\times224$ \\
         PLGMI \citep{plg} & \textit{White-box} & AAAI-2023 & Conditional GAN & $64\times64$ \\
         LOMMA \citep{lomma} & \textit{White-box} & CVPR-2023 & $\sim$ & $64\times64$ \\
         IF-GMI \citep{ifgmi} & \textit{White-box} & ECCV-2024 & StyleGAN2 & $224\times224$ \\
         \midrule
         Mirror$^*$ \citep{mirror} & \textit{White-box/Black-box} & NDSS-2022 & StyleGAN & $224\times224$ \\
         C2FMI \citep{c2f} & \textit{Black-box} & TDSC-2023 & StyleGAN2& $160\times160$ \\
         RLBMI \citep{rlb} & \textit{Black-box} & CVPR-2023 & WGAN & $64\times64$ \\
         \midrule
         BREPMI \citep{brep} & \textit{Label-only} & CVPR-2022 &  WGAN & $64\times64$ \\
         LOKT \citep{lokt} & \textit{Label-only} & NeurIPS-2023 & ACGAN & $128\times128$ \\
        \bottomrule
    \end{tabular}
    }
    \begin{tablenotes}[flushleft]
	\item { $^*$Mirror \citep{mirror} proposes attack methods on both \textit{white-box} and \textit{black-box} settings.}
        \item { $^{\dag}$KEDMI \citep{ked} first proposed this customized GAN.}
        \item { $^{\sim}$LOMMA \citep{lomma} is a plug-and-play technique applied in combination with other MI attacks.}
     \end{tablenotes} 
    \end{threeparttable}
\end{table*}

\begin{table*}[!ht]
    \setlength{\tabcolsep}{5pt}
    \normalsize
    \centering
    \caption{Summary of implemented MI defense strategies in our benchmark.}
    \label{table:defenses}
    \begin{threeparttable} 
    \resizebox{\linewidth}{!}{
    \begin{tabular}{ccccc}
        \toprule
         \textbf{Defense Strategy} & \textbf{Functional Stage} & \textbf{Reference} & \textbf{Core Technique} & \textbf{Description}\\ \midrule
         MID \citep{mid} & Train-time & AAAI-2021 & Regularization & \makecell[l]{Utilize mutual information regularization to limit leaked information \\ about the model input in the model output} \\ \midrule
         BiDO \citep{bido} & Train-time & KDD-2022 & Regularization & \makecell[l]{Minimize dependency between latent vectors and the model input \\ while maximizing dependency between latent vectors and the outputs} \\ \midrule
         LS \citep{ls} & Train-time & ICLR-2024 & Label Smoothing & \makecell[l]{Adjusting the label smoothing with negative factors contributes to \\ increasing privacy protection} \\ \midrule
         TL \citep{tl} & Train-time & CVPR-2024 & Transfer Learning & \makecell[l]{Utilize transfer learning to limit the number of layers encoding \\ privacy-sensitive information for robustness to MI attacks} \\ \midrule
         RoLSS/SSF/TTS$^*$ \citep{TTS} & Train-time & ECCV-2024 & Skip Connection & \makecell[l]{Properly remove some skip connection structures in the target classifier \\ during the training stage} \\ \midrule
         Purifier \citep{yang2023purifier} & Inference-time & AAAI-2023 & Conditional VAE & \makecell[l]{Reform confidences and swap labels to make members behave \\ like non-members} \\ \midrule
         SSD \citep{zhuangstealthy} & Inference-time & ICLR-2025 & \makecell[c]{Conditional\\Mutual Information} & \makecell[l]{Minimize conditional mutual information to make predictions\\less dependent on inputs and more dependent on ground truths} \\
        \bottomrule
    \end{tabular}
    }
    \begin{tablenotes}[flushleft]
	\item { $^*$The three defenses proposed by \citep{TTS} have minor distinction between each other, therefore we select the TTS as the \\ representative method in our evaluation.}
     \end{tablenotes} 
    \end{threeparttable}
\end{table*}
\section{Our Benchmark}

\subsection{Dataset}

Considering existing MI attacks primarily focus on reconstructing private facial data from image classifiers, we select 4 widely recognized face datasets as the basic datasets, including Flickr-Faces-HQ (FFHQ) \citep{karras2019style}, MetFaces \citep{metfaces}, FaceScrub \citep{facescrub}, and CelebFaces Attributes (CelebA) \citep{celeba}. Generally, FFHQ and MetFaces are employed as public datasets for pre-training auxiliary priors, whereas FaceScrub and CelebA serve as the target private datasets to attack. Our benchmark is modularized to facilitate researchers to freely combine public datasets with private ones, enabling customized experimental setups. Extensive evaluation on more non-facial datasets (Stanford Dogs \citep{standforddog} and AFHQ-Dog \citep{afhq}) are presented in Supp.\ref{combination}.

Notably, the target resolutions across different MI attacks are not uniform. The majority of attack methods concentrate on low-resolution images of $64\times64$, while recent attack methods have begun to focus on higher resolutions, such as $224\times224$. Therefore, our benchmark offers 2 versions of low-resolution and higher-resolution for the aforementioned 4 datasets and prepares multiple transformation tools for processing images, freeing researchers from the laborious tasks of data preprocessing. More details regarding the datasets can be found in the Supp.\ref{appendix:dataset}.

\subsection{Implemented Methods}

Our benchmark includes a total of 19 methods, comprising 12 attack methods and 7 defense strategies. With a focus on Generative Adversarial Network (GAN)-based MI attacks, we selectively reproduce methods from recent years that have been published in top-tier conferences or journals in the computer vision or machine learning domains. This criterion ensures the reliability and validity of the implemented methods. Considering the main targets in our benchmark are image classifiers for RGB images, the learning-based MI attacks \citep{fredrikson2015model, song2017machine, yang2019neural} are not incorporated currently. More detailed information about the implemented methods is stated in Supp.\ref{appendix:attack} and Supp.\ref{Benchmark Details-Details of Classifier Training}.

\textbf{\textit{Attacks.}} Based on the accessibility to the target model's parameters, we categorize MI attacks into \textit{white-box} and \textit{black-box} attacks. \textit{White-box} attacks can entail full knowledge of the target model, enabling the computation of gradients for performing backpropagation, while \textit{black-box} attacks are constrained to merely obtaining the prediction confidence vectors of the target model. Our benchmark includes 8 \textit{white-box} attack methods and 4 \textit{black-box} attack methods, as summarized in Table \ref{table:attacks}.

\textbf{\textit{Defenses.}} To effectively defend against MI attacks, the defender typically employs defense strategies during the training process of victim classifiers. Our benchmark has implemented 4 typical defense strategies and corresponding details are presented in Table \ref{table:defenses}.

\begin{figure*}[tbp]
\centerline{\includegraphics[width=1.8\columnwidth]{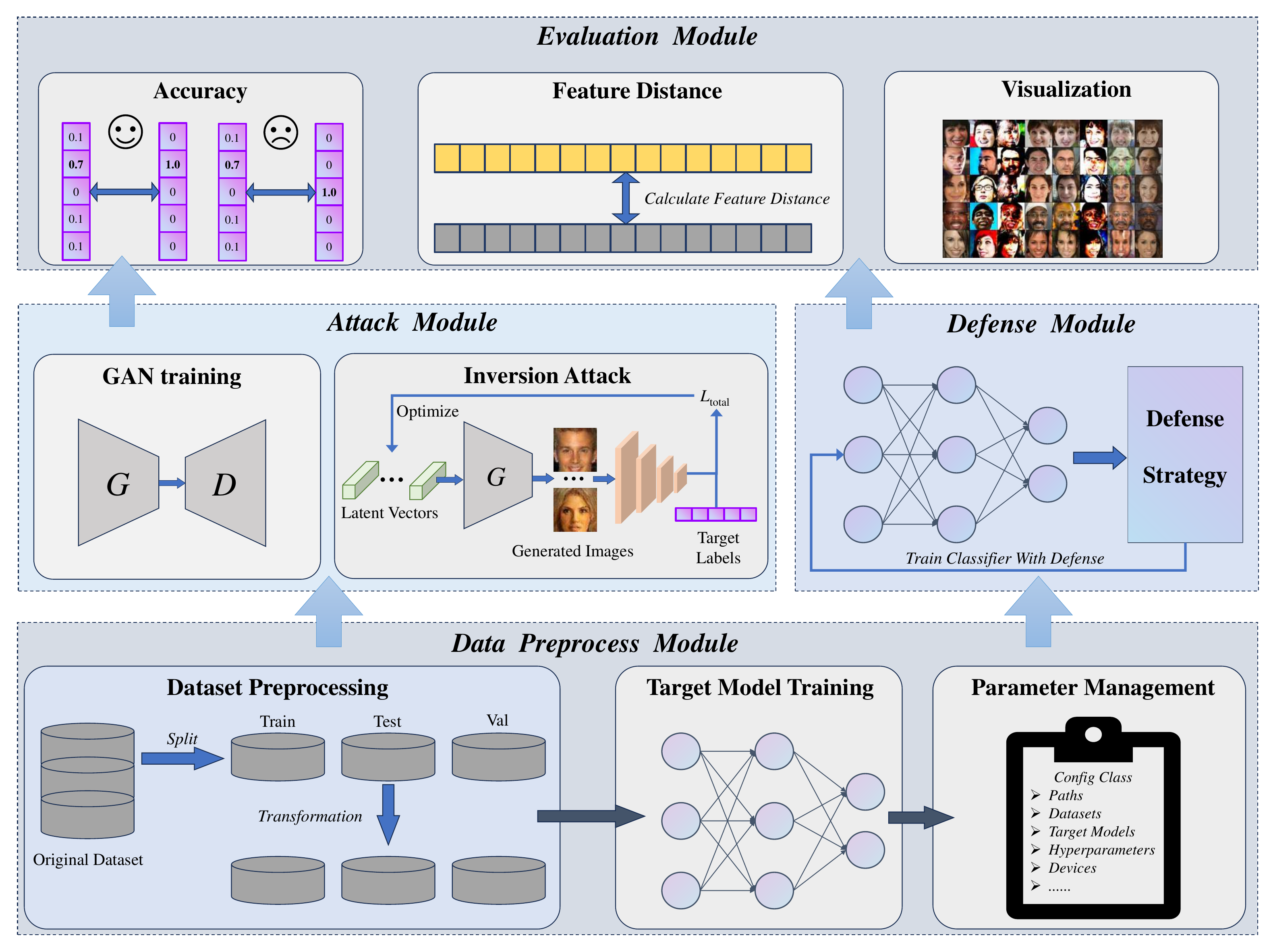}}
\caption{Overview of the basic structure of modular-based toolbox for our benchmark.}
\label{overview}
\end{figure*}

\subsection{Toolbox}

We implement an extensible and reproducible modular-based toolbox for our benchmark, as shown in Fig. \ref{overview}. The framework can be roughly divided into four primary modules, including \textit{Data Preprocess Module}, \textit{Attack Module}, \textit{Defense Module} and \textit{Evaluation Module}.

\textit{\textbf{Data Preprocess Module.}} This module is designed to preprocess all data resources required before launching attacks or defenses, including datasets, classifiers and parameters. Consequently, we furnish this module with three fundamental functionalities: \textit{dataset preprocessing}, \textit{target model training}, and \textit{parameter management}. For \textbf{dataset preprocessing}, we build a unified pipeline for each dataset, which automatically carries out a series of operations such as splitting dataset and image transformations (\textit{e.g.} center crop and resize) based on the split file and chosen resolution from users. For \textbf{target model training}, users can further leverage the processed datasets to train designated classifiers. We abstract the general procedures in classifier training into a base trainer class to facilitate users in extending and customizing their own classifiers. For \textbf{parameter management}, we encapsulate parameters used in different processes into specific configuration classes, such as TrainConfig, designated for the training process, and AttackConfig for the attack process, thus ensuring organized and efficient parameter handling in the workflow.

\textit{\textbf{Attack Module.}} The workflow of MI attacks can be roughly divided into two stages. The first stage is \textbf{GAN training}, where the module abstracts the general training process of GANs into a basic trainer class. This allows users not only train GANs that are pre-built into the benchmark, but also extend to their uniquely designed GANs. The second stage is the core \textbf{inversion attack}, which we split into three parts: \textit{latent vectors initialization}, \textit{iterative optimization}, and an optional \textit{post-processing} step. After the completion of the attack, the module preserves essential data such as the final optimized images and latent vectors, facilitating subsequent evaluation and analysis.

\textit{\textbf{Defense Module.}} Considering the mainstream MI defense strategies are applied during the training process of target classifiers, we design the defense module following the target model training functionality within the \textit{Data Preprocessing Module}. To enhance extensibility, we incorporate defense strategies as part of the training parameters for classifiers to enable the defense during the training process of target models, which decouples the defense from the training pipeline. In this way, we allow users to customize their own defense strategies against MI attacks.

\textit{\textbf{Evaluation Module.}} Following the common setting in MI field, our benchmark concentrates on the evaluation at distribution level instead of sample level, assessing the overall performance of the whole reconstructed dataset. Therefore, we provide a total of 9 widely recognized distribution level evaluation metrics for users, which can be categorized into four types according to the evaluated content:
\label{Section:benchmark: Details of Evaluation}
\begin{itemize}
    \item \textbf{Attack Accuracy.} This metric serves as the primary measurement for MI attacks, using a given classifier to classify the inverted images and measures the top-$1$ ($\mathbf{Acc}@1$) and top-$5$ ($\mathbf{Acc}@5$) accuracy for target labels. The higher the reconstructed samples achieve attack accuracy on another classifier trained with the same dataset, the more private information in the dataset can be considered to be exposed \citep{gmi}. Therefore, higher attack accuracy denotes better attack performance. 
    \item \textbf{Feature Distance.} The feature is defined as the output of the given classifier's penultimate layer. We compute the shortest feature $l_2$ distance between inverted images and private samples for each class and calculate the average distance. In our benchmark, we follow the common choices across previous methods \cite{ppa,ifgmi} and select the evaluation model and another pre-trained FaceNet \cite{schroff2015facenet} to compute the feature distance, respectively denoted as $\delta_{eval}$ and $\delta_{face}$. Smaller feature distance means more similar features to the private image.
    \item \textbf{Fr\'echet Inception Distance (FID).} FID \citep{fid} is commonly used to evaluate the generation quality of GANs by computing the distance between feature vectors from all the target private data and that from all the reconstructed samples. Feature vectors are extracted by an Inception-v3 \cite{inceptionv3} pre-trained on ImageNet \cite{imagenet}. Lower FID score shows higher inter-class diversity and realism \citep{vmi}.
    \item \textbf{Sample Diversity.} The metric contains Precision-Recall \citep{pr} and Density-Coverage \citep{dc} scores. Higher values indicate greater intra-class diversity of the inverted images. Since this metric serves as a secondary indicator for auxiliary reference in MI tasks, we have placed the corresponding evaluation results in the Supp.\ref{prdc}.
\end{itemize}
\section{Evaluation}
\label{experiment}
\subsection{Evaluation Setups}

To ensure fair and uniform comparison and evaluation, we select the FFHQ \citep{karras2019style} as the public dataset and FaceScrub \citep{facescrub} as the private dataset for all the experiments in the Experiment section. The target models are fixed to the IR-152 \citep{resnet} for low-resolution scenario and ResNet-152 \citep{resnet} for high-resolution scenario, both trained on the FaceScrub. For each attack method, the number of images reconstructed per class is set to 5 due to time and computation cost. More detailed experimental settings are listed in Supp.\ref{B}.

Notably, we limit the evaluation exhibited in the Experiment section to merely three metrics, including \textit{Accuracy}, \textit{Feature Distance}, and \textit{FID} \citep{fid}, while \textit{Sample Diversity} is presented in the Supp.\ref{prdc} due to the paper length limitation. To measure the deviation across the target labels, we introduce error bars for \textit{Accuracy} and \textit{Feature Distance}. As \textit{FID} metric is computed based on all the recovered images, it is not essential to calculate error bars. Moreover, VMI \citep{vmi} and RLBMI \citep{rlb} will be further evaluated in Supp. \ref{v-r-p} owing to their excessive need of time (See Supp. \ref{appendix: time}). We also conduct a user study in Supp. \ref{appendix: user-study}.

\begin{figure*}[tbp]
\centerline{\includegraphics[width=\textwidth]{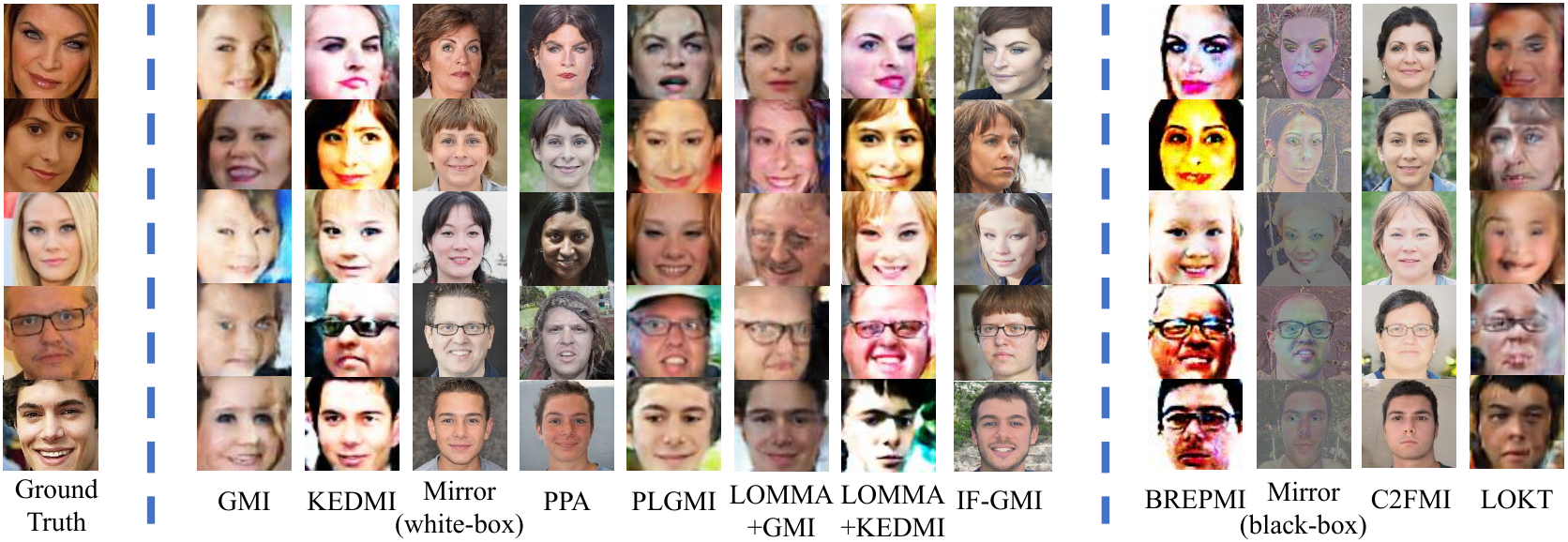}}
\caption{Visual comparison between different MI attacks. As MI attacks are designed to reconstruct private features of target images at the distribution level rather than strictly full reconstruction of original images, it is sufficient for the inverted images to exhibit discriminative characteristics associated with the original image category.}
\label{visual-low}
\end{figure*}

\subsection{Evaluation on Different Attack Methods}
\label{section:Evaluation on Different Attack Methods}

In this part, we prepare a unified experimental setting for different MI attack methods to ensure fair comparison. The resolution of private and public datasets is set to $64\times64$, indicating relatively easier scenarios. Comparisons of white-box and black-box MI attacks are presented in Table \ref{white:different methods}. 

Remarkably, the PLGMI \citep{plg} and LOKT \citep{lokt} achieve state-of-the-art comprehensive performance in white-box attacks and black-box attacks respectively, showing significant superiority in \textit{Accuracy} and \textit{Feature Distance} metrics. However, the lowest \textit{FID} scores occur in the PPA \citep{ppa} and C2FMI \citep{c2f} respectively instead of the above methods. We infer that this is because PPA and C2FMI employ more powerful generators (\textit{e.g.} StyleGAN2 \citep{stylegan2}) as the GAN prior compared to PLGMI and LOKT, leading to more real image generation. Visualization results displayed in Fig \ref{visual-low} further validate the inference. 

Notably, the visualization of Mirror(black) in Fig.\ref{visual-low} and Fig.\ref{visual-high} (in Supp.\ref{appendix: visualization}) is anomalous compared to other methods. We perform further investigation on this special phenomenon in Supp.\ref{appendix: mirror-visual}. The experimental results demonstrate that this phenomenon is highly likely attributable to the genetic algorithm compromising the integrity of StyleGAN's $\mathcal{W}$ latent vectors, consequently inducing a deviation in image generation from the normal distribution.

\begin{table}[!ht]
    \setlength{\tabcolsep}{5pt}
    \normalsize
    \centering
    \caption{Comparison between different white-box and black-box MI attacks. The value after $\pm$ indicates the error bar across all the target labels.}
    \label{white:different methods}
    \begin{threeparttable} 
    \resizebox{\linewidth}{!}{
    \begin{tabular}{ccccccc}
        \toprule
\textbf{Method} & $\uparrow{\mathbf{Acc}@1}$ & $\uparrow{\mathbf{Acc}@5}$ & $\downarrow\mathbf{\delta}_{eval}$ & $\downarrow\mathbf{\delta}_{face}$ & $\downarrow\textbf{FID}$\\ \midrule
GMI & $0.153\pm0.077$ & $0.265\pm0.093$ & $2442.667\pm298.597$ & $1.300\pm0.176$ & $91.861$\\
KEDMI & $0.404\pm0.017$ & $0.579\pm0.013$ & $2113.473\pm545.085$ & $0.997\pm0.337$ & $61.035$\\
Mirror(white) & $0.311\pm0.014$ & $0.509\pm0.021$ & $1979.211\pm427.343$ & $0.996\pm0.258$ & $36.610$\\
PPA & $0.844\pm0.036$ & $0.923\pm0.026$ & $1374.967\pm387.380$ & $0.657\pm0.195$ & $\mathbf{31.433}$\\
PLGMI & $\mathbf{0.998}\pm0.002$ & $\mathbf{0.999}\pm0.001$ & $\mathbf{967.295}\pm222.725$ & $\mathbf{0.486}\pm0.103$ & $74.155$\\
LOMMA+GMI & $0.557\pm0.111$ & $0.678\pm0.096$ & $1948.976\pm317.310$ & $0.949\pm0.221$ & $62.050$\\
LOMMA+KEDMI & $0.711\pm0.007$ & $0.860\pm0.006$ & $1685.514\pm486.419$ & $0.759\pm0.289$ & $62.465$\\
IF-GMI & $0.797 \pm 0.018$ & $0.865 \pm 0.014$ & $1462.914\pm 486.419$ & $0.722\pm 0.232$ & $33.057$ \\
\midrule
Mirror(black) & $\textbf{0.526}\pm0.031$ & $\textbf{0.729}\pm0.020$ & $\textbf{1972.175}\pm427.391$ & $\textbf{0.854}\pm0.239$ & $54.231$\\
C2FMI & $0.263\pm0.009$ & $0.459\pm0.016$ & $2061.995\pm534.556$ & $1.011\pm0.265$ & $\textbf{43.488}$\\
\midrule
BREPMI & $0.354\pm0.013$ & $0.608\pm0.015$ & $2178.587\pm357.194$ & $0.971\pm0.186$ & $74.519$\\
LOKT & $\textbf{0.834}\pm0.010$ & $\textbf{0.918}\pm0.013$ & $\textbf{1533.071}\pm402.791$ & $\textbf{0.694}\pm0.169$ & $\textbf{71.701}$\\
\bottomrule
    \end{tabular}
    }
    \end{threeparttable}
\end{table}

\subsection{Evaluation on Higher Resolution}
\label{section:Evaluation on Higher Resolution}

Recent attack methods have attempted to conquer higher resolution scenarios, such as PPA and Mirror. Accordingly, we conduct a further assessment of MI attacks under an increased resolution of $224\times224$, which is considered a more challenging task. The evaluation results for white-box and black-box attacks are demonstrated in Table \ref{white:high resolution}. 

The results imply the significant impact of GAN priors when attacking private images with higher resolution. All the methods that employ stronger GAN priors maintain low FID scores, including Mirror, C2FMI, and PPA, while other methods suffer from significant degradation in image reality. Despite the primary metric for evaluating MI attacks is \textit{Accuracy}, the reality of reconstructed images should be ensured within a reasonable range for better image quality. Thus, it is imperative to explore more complex GAN priors with enhanced performance in future research, extending the MI field to more challenging and practical applications. More corresponding visualization results are displayed in Supp.\ref{appendix: visualization}.

\begin{table}[!ht]
    \setlength{\tabcolsep}{5pt}
    \normalsize
    \centering
    \caption{Comparison between MI attacks on higher resolution scenario.}
    \label{white:high resolution}
    \begin{threeparttable} 
    \resizebox{\linewidth}{!}{
    \begin{tabular}{ccccccc}
        \toprule
\textbf{Method} & $\uparrow{\mathbf{Acc}@1}$ & $\uparrow{\mathbf{Acc}@5}$ & $\downarrow\mathbf{\delta}_{eval}$ & $\downarrow\mathbf{\delta}_{face}$ & $\downarrow\textbf{FID}$\\ \midrule
GMI & $0.073\pm0.024$ & $0.192\pm0.056$ & $\textbf{134.640}\pm24.203$ & $1.328\pm0.135$ & $119.755$\\
KEDMI & $0.252\pm0.007$ & $0.494\pm0.013$ & $144.139\pm33.673$ & $1.139\pm0.214$ & $124.526$\\
Mirror(white) & $0.348\pm0.023$ & $0.649\pm0.016$ & $197.741\pm32.212$ & $1.049\pm0.154$ & $59.628$\\
PPA & $0.913\pm0.022$ & $0.986\pm0.004$ & $167.532\pm28.944$ & $0.774\pm0.143$ & $\mathbf{46.246}$\\
PLGMI & $\mathbf{0.926}\pm0.007$ & $\mathbf{0.987}\pm0.002$ & $135.557\pm36.500$ & $0.730\pm0.177$ & $117.850$\\
LOMMA+GMI & $0.735\pm0.043$ & $0.875\pm0.037$ & $136.700\pm29.743$ & $0.953\pm0.171$ & $111.151$\\
LOMMA+KEDMI & $0.627\pm0.009$ & $0.864\pm0.006$ & $146.612\pm42.594$ & $0.977\pm0.244$ & $103.479$\\ 
IF-GMI & $0.815\pm0.015$ & $0.958\pm 0.003$  &$263.081\pm 62.775$ & $\mathbf{0.711}\pm 0.146$ &
$47.590$ \\ \midrule
Mirror (black) & $\textbf{0.611}\pm0.051$ & $\textbf{0.862}\pm0.018$ & $\textbf{198.609}\pm40.255$ & $\textbf{1.049}\pm0.192$ & $92.413$\\
C2FMI & $0.414\pm0.017$ & $0.686\pm0.018$ & $439.659\pm93.688$ & $1.592\pm0.249$ & $\textbf{47.317}$\\ \midrule
BREPMI & $\textbf{0.342}\pm0.013$ & $\textbf{0.622}\pm0.026$ & $134.263\pm31.441$ & $\textbf{1.067}\pm0.208$ & $\textbf{105.489}$\\
LOKT & $0.328\pm0.004$ & $0.553\pm0.010$ & $\mathbf{126.964}\pm36.434$ & $1.122\pm0.284$ & $127.709$\\
\bottomrule
    \end{tabular}
    }
    \end{threeparttable}
\end{table}

\subsection{Evaluation on Varied Model Predictive Power}

The predictive power of the target model is a crucial factor in determining the effectiveness of MI attacks. Previous work \citep{gmi} has conducted preliminary experimental validations on simple networks (\textit{e.g.} LeNet \citep{lecun1989handwritten}), demonstrating that the performance of the first GAN-based attack GMI is influenced by the predictive power of the target model. Therefore, we evaluate the state-of-the-art MI attacks on target models with varied predictive power to further validate the consistency of this characteristic in the recent attacks. The resolution of datasets is set to $64\times64$ and the evaluation results are shown in Fig. \ref{predictive}.

The comparison in Fig. \ref{predictive} reveals most MI attacks maintain the trend that higher predictive power contributes to better attack performance, which is consistent with the aforementioned characteristic. Specifically, the earlier attack methods (\textit{e.g.}, GMI and KEDMI) presents more fluctuation on the trend across different predictive power, while the recent attack methods (\textit{e.g.}, PLGMI and PPA) show in more stable trend when predictive power increases. This indicates the predictive power of target models plays a crucial role in measuring the performance of MI attacks. Thus one can expect lower privacy leakage in the robust model by balancing the accuracy-privacy trade-off.

\begin{figure*}[tbp]
\centerline{\includegraphics[width=\linewidth]{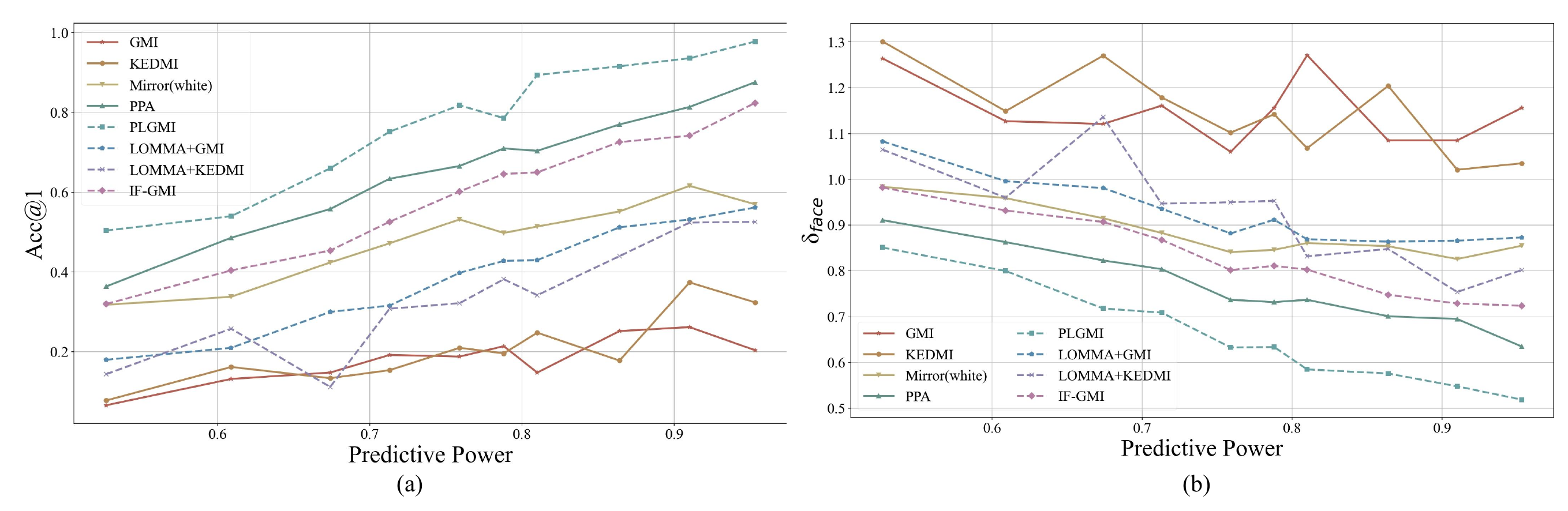}}
\caption{Comparison across ResNet-152 with varied predictive power. (a) The incremental trend of $Acc@1$ metric on different attack methods. (b) The decreasing trend of $\delta_{face}$ metric on different attack methods.}
\label{predictive}
\end{figure*}


\begin{figure}[tbp]
\centerline{\includegraphics[width=\columnwidth]{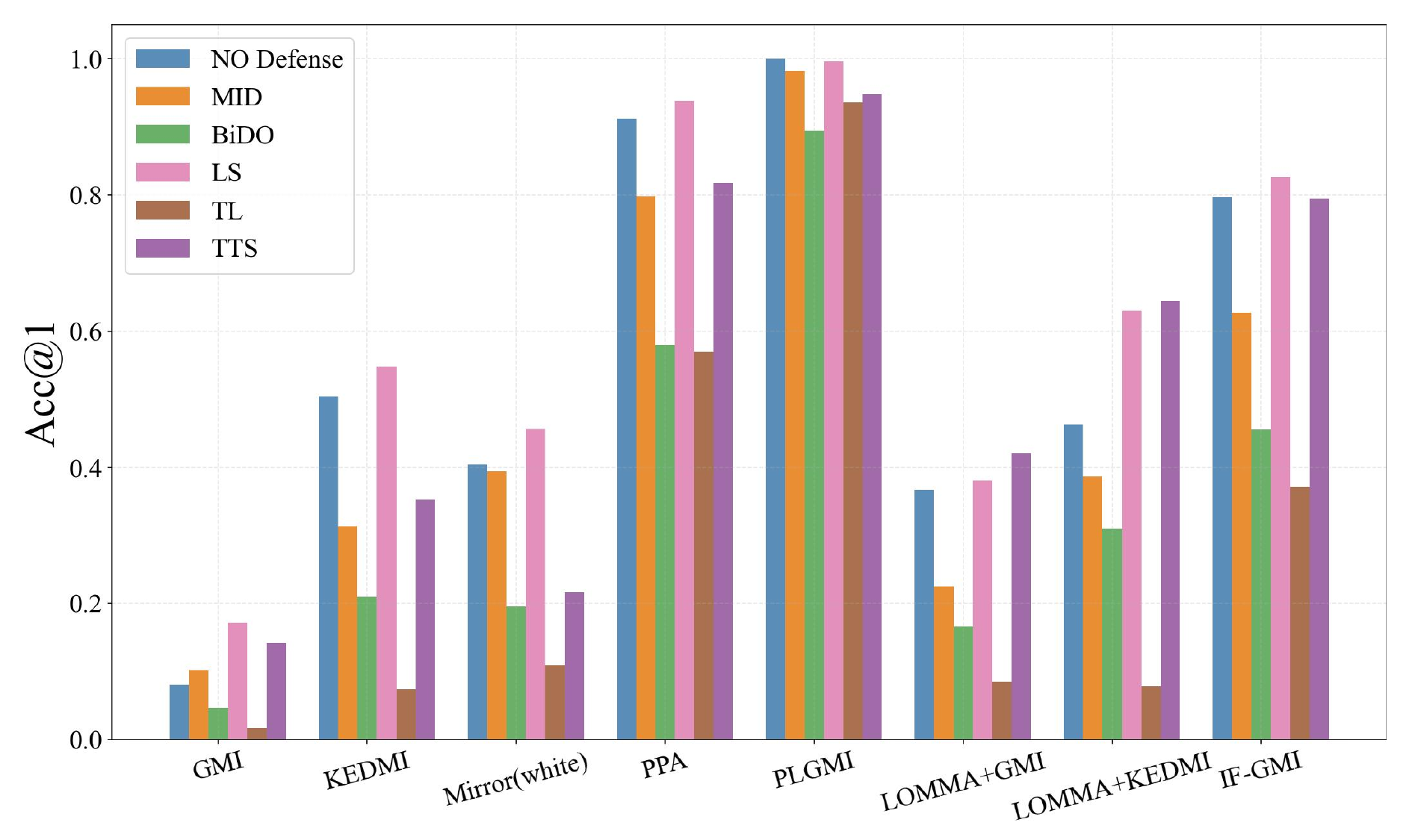}}
\caption{Evaluation of Different white-box MI attacks on multiple MI defense strategies. Notably, attackers can obtain full access to the model parameters under the white-box settings, enabling them to directly circumvent inference-time defense plugins. Consequently, such defenses are invalid to white-box attacks.}
\label{train-time defense}
\end{figure}

\begin{figure}[tbp]
\centerline{\includegraphics[width=\columnwidth]{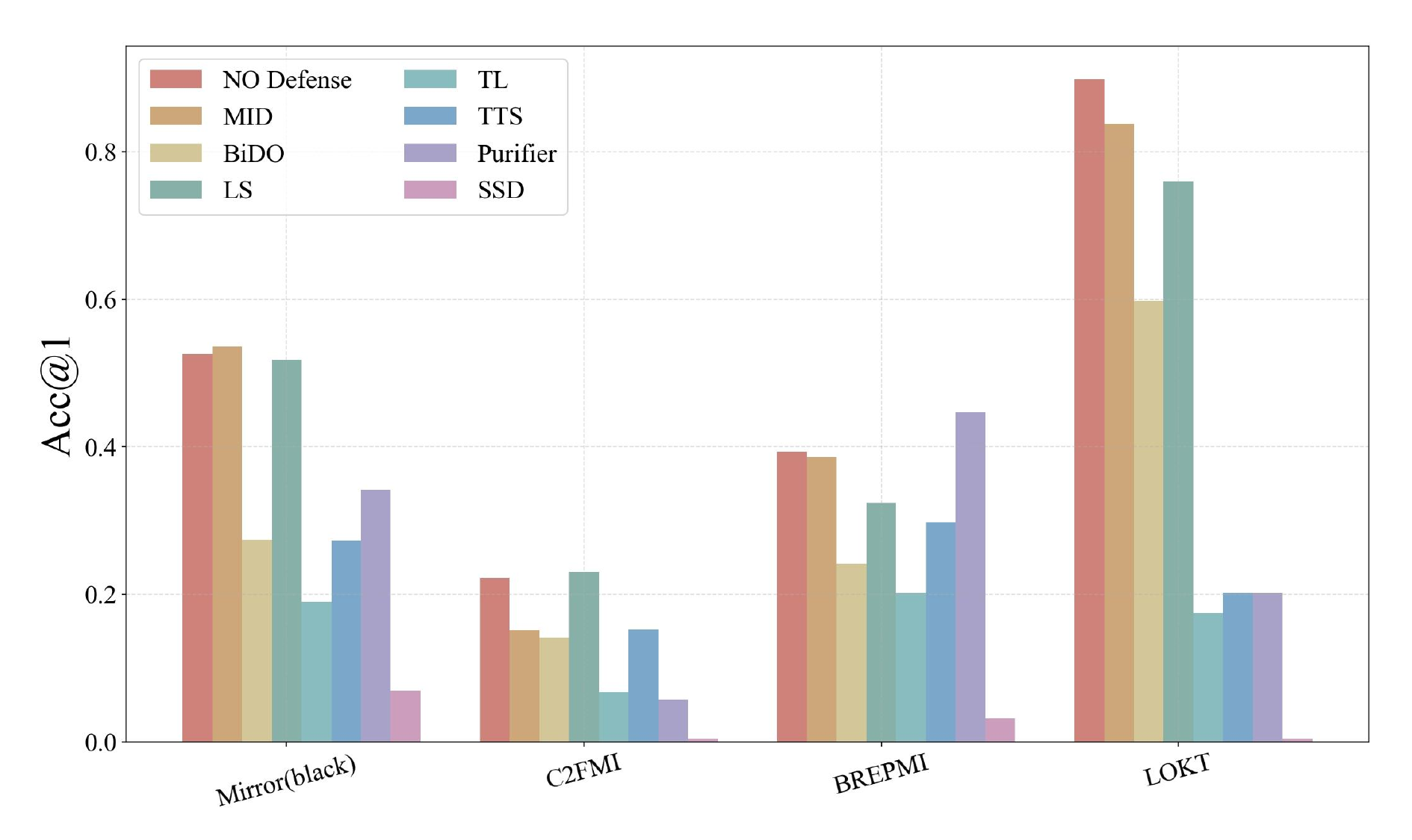}}
\caption{Evaluation of Different black-box and label-only MI attacks on multiple MI defense strategies.}
\label{inference-time defense}
\end{figure}

\subsection{Evaluation on Defense Strategies}
\label{section:Evaluation on Defense Strategies}

This analysis focuses on the robustness of MI attacks when applied to models with defenses. Notably, we select the first 100 classes subset from FaceScrub as the target dataset due to the time cost. The assessment results are listed in Fig.\ref{train-time defense} and Fig.\ref{inference-time defense}. The configuration of defenses is set following the official parameters, as detailed in Table \ref{table:classifier-defense} in the Supplementary.

From Fig.\ref{train-time defense}, the TL \citep{tl} achieves the state-of-the-art decrease in \textit{Accuracy}. However, advanced attack methods have overcome current defense strategies to some extent, such as PLG and PPA. Additionally, some older defense strategies (\textit{e.g.} MID \citep{mid}) are no longer effective against the latest attacks. As for Table \ref{inference-time defense}, the SSD \citep{zhuangstealthy} is especially effective when defending against black-box and label-only attacks. This indicates that employing defense in the inference stage is more effective than in the training process.

From Fig.\ref{train-time defense}, we observe that LS \citep{ls} exhibits unexpectedly poor performance in $\mathbf{Acc}@1$ metric while it was recently published in the top-tier conference. The potential reason might be the utilized target classifier with relatively high test accuracy, as validated in the Supp.\ref{appendix: ls}. Furthermore, we conduct further experiments with PPA as the attack method against ResNet-152 trained on high resolution (224×224) scenario, proving that this phenomenon can be extended into other defenses. The results are listed in the Table \ref{table:defense_extend}, demonstrating that all the train-time defenses become invalid even with the recommended parameters. More in-depth evaluation on defenses is exhibited in Supp.\ref{appendix: low-acc}.

Combining this phenomenon with the above experiment on predictive power, our empirical analysis indicates that the leaked information is strongly correlated to the model prediction accuracy, and current defenses cannot effectively reduce the privacy information without sacrificing of model performance. Our findings emphasize that more reliable and stable defense strategies should be studied. A potential solution is designinga  more robust model architecture instead of merely adding a regularizer in the training process.

\subsection{Evaluation on Adversarial Robustness}

Most defense methods are designed to focus solely on countering MI attacks, while neglecting protection against other types of attacks. However, since both training-time and inference-time defenses can influence the model's outputs, it becomes imperative to evaluate the robustness of defense-protected models when confronted with adversarial attacks. This necessity highlights the importance of comprehensive security assessments to ensure that enhanced privacy protections do not inadvertently compromise the model's resilience against diverse attacks.

We select the typical adversarial attacks FGSM \citep{fgsm}, PGD \citep{pgd}, BIM \citep{bim}, and OnePixel \citep{onepixel} as representative to evaluate different MI defenses. The target models are set to IR-$152$ trained under multiple defenses on the high-resolution FaceScrub dataset. The results are shown in Table \ref{tab:adversarial}. In general, the inference-time defenses demonstrate extraordinary enhancement in adversarial robustness, while train-time defenses fail to improve adversarial robustness. Especially, most train-time defenses reduce adversarial robustness under the untargeted scenario of OnePixel.  

Our findings reveal that train-time defenses fail to adequately address multi-faceted robustness requirements. A promising solution is integrating adversarial training into the train-time defense process to enhance robustness.

\begin{table}[!ht]
    \setlength{\tabcolsep}{5pt}
    \normalsize
    \centering
    \caption{Extensive evaluation on multiple MI defense strategies.}
    \label{table:defense_extend}
    \begin{threeparttable} 
    \resizebox{\linewidth}{!}{
    \begin{tabular}{cccccccc}
        \toprule
\textbf{Method} & \textbf{Hyperparameters} & \textbf{Test Acc} & $\uparrow{\mathbf{Acc}@1}$ & $\uparrow{\mathbf{Acc}@5}$ & $\downarrow\mathbf{\delta}_{eval}$ & $\downarrow\mathbf{\delta}_{face}$ & $\downarrow\textbf{FID}$\\ \midrule
NO Defense & - & 0.985 & 0.972 & 0.990 & 307.714 & 0.588 & 50.259 \\ \midrule
\multirow{2}{*}{MID} & $\alpha=0.005$ & 0.967 & 1.000 & 1.000 & 273.687 & 0.517 & 49.239 \\
~ & $\alpha=0.01$ & 0.956 & 0.990 & 1.000 & 276.889 & 0.526 & 51.227 \\ \midrule
\multirow{2}{*}{BiDO} & $\alpha=0.01,\beta= 0.1$ & 0.980 & 0.986 & 0.996 & 306.827 & 0.554 & 50.943 \\
~ & $\alpha=0.05,\beta= 0.5$ & 0.974 & 0.968 & 0.996 & 320.191 & 0.594 & 50.132 \\ \midrule
\multirow{2}{*}{LS} & $\alpha=-0.005$ &0.967 &  0.971 & 0.996 & 281.956 & 0.524 & 38.924\\
& $\alpha=-0.01$ & 0.964 & 0.949 & 0.995 & 302.442& 0.563 & 39.185 \\\midrule
\multirow{2}{*}{TL} & $\alpha=0.4$ & 0.976 & 0.994 & 1.000 & 282.394 & 0.513 & 53.023 \\
~ & $\alpha=0.5$ & 0.975 & 0.982 & 0.996 & 306.528 & 0.561 & 51.489 \\\midrule
\multirow{2}{*}{TTS} & $\alpha=0,n=1$ &0.959 & 0.861 & 0.967 & 143.637 & 	0.706 & 	41.243\\
 & $\alpha=0.01,n=1$ & 0.968 & 0.857	 & 0.968 & 	147.672	 & 0.707	 & 42.002\\
        \bottomrule
    \end{tabular}
    }
    \end{threeparttable}
\end{table}

\begin{table}[!ht]
    \setlength{\tabcolsep}{5pt}
    \normalsize
    \centering
    \caption{Adversarial attack results against IR-$152$ models trained on FaceScrub dataset in the low-resolution scenario. The evaluation metric is Attack Success Rate (ASR).}
    \label{tab:adversarial}
    \begin{threeparttable} 
    \resizebox{\linewidth}{!}{
    \begin{tabular}{cccccccccc}
        \toprule
 \multirow{2}{*}{\textbf{Method}} 
 &  \multirow{2}{*}{\textbf{Test Acc}} 
 & \multicolumn{4}{c}{Untargeted Attacks} &\multicolumn{4}{c}{Targeted Attacks} \\
 \cmidrule(lr){3-6}\cmidrule(lr){7-10} &  &
  $\downarrow$ \textbf{FGSM} & $\downarrow$ \textbf{PGD} & $\downarrow$ \textbf{BIM} & $\downarrow$ \textbf{OnePixel} & $\downarrow$ \textbf{FGSM} & $\downarrow$ \textbf{PGD} & $\downarrow$ \textbf{BIM} & $\downarrow$ \textbf{OnePixel} \\\midrule

$\textbf{No defense}$ & 0.982& 0.997& 0.995& 1.000& 0.665& 0.087& 0.170& 0.161 & 0.000\\ \midrule
$\textbf{MID}$ & 0.982& 1.000& 0.997& 0.997& 0.695& 0.087& 0.139 & 0.153& 0.002\\
$\textbf{BiDO}$ & 0.940& 1.000& 1.000& 1.000& 0.826& 0.085& 0.114& 0.112& 0.007\\
$\textbf{LS}$ & 0.980& 0.997& 0.997& 0.997& 0.665& 0.092& 0.158& 0.148& 0.007\\
$\textbf{TL}$ & 0.953& 1.000& 1.000& 1.000& 0.873& 0.117& 0.175& 0.170& 0.007\\
$\textbf{TTS}$ & 0.973& \textbf{0.826}& 0.985& 0.992& 0.100& 0.034& 0.497& 0.502& \textbf{0.000}\\
$\textbf{Purifier}$ & 0.946& 0.939& \textbf{0.897}& \textbf{0.934}& \textbf{0.075}& \textbf{0.002}& \textbf{0.060}& 0.068& \textbf{0.000}\\
$\textbf{SSD}$ & 0.950& 0.902& 0.995& 0.987& 0.209& 0.004& 0.087& \textbf{0.061} & \textbf{0.000}\\

\bottomrule
    \end{tabular}
    }
    \end{threeparttable}
\end{table}
\section{Conclusion}

In this paper, we develop \textit{MIBench}, a comprehensive, unified, and reliable benchmark, and provide an extensible and reproducible toolbox for researchers. To the best of our knowledge, this is the first benchmark and first open-source toolbox in the MI field. Our benchmark encompasses 19 of the state-of-the-art MI attack methods and defense strategies, and more algorithms will be continually updated. Based on the implemented toolbox, we establish a consistent experimental environment and conducted extensive evaluation to facilitate fair comparison between different methods. With in-depth analyses, we have benchmarked all the typical MI attacks and defenses under the unified setups and identified new insights for future study.

\label{societal}
\textbf{Societal Impact and Ethical Considerations.} A potential negative impact could be malicious users leveraging the implemented attack methods to reconstruct private data from the public system. To alleviate potential dilemma, a cautious approach is to establish access permissions.


\newpage
\appendix
\endgroup
\setcounter{tocdepth}{2}  

\begingroup
\renewcommand{\contentsname}{Appendix Table} 
\tableofcontents  
\endgroup

\newpage
\newpage

\section{Benchmark Details}

\subsection{Details of Data Processing}

This section introduces three types of dataset processing implemented by our toolbox, including data pre-processing, dataset splitting, and dataset synthesis. 

\subsubsection{Data Pre-processing.}
\label{Benchmark Details Data Pro-processing}
Data pre-processing aligns face images to ensure consistency across datasets. Users can customize the transformations to be used for data pre-processing. 
We also provide default processing for four datasets in high and low versions of the resolution. 
Low resolution versions include $64\times64$ and $112\times112$, and high resolution versions include $224\times224$ and $299\times299$.
Here are the default pre-processing method for the four datasets.

\begin{itemize}
    \item \textbf{CelebA \citep{celeba}.} For the low version, a center-croped with a crop size $108\times 108$ and a resize function is applied to each sample from the origin images. For the high version, only a direct resize will be used in the images pre-processed from HD-CelebA-Cropper \citep{hdcelebacropper}.
    \item \textbf{FaceScrub \citep{facescrub}.} In the low-resolution version, the original data is center cropped at $54/64$ and then scaled, and the high-resolution is scaled directly.
    \item \textbf{FFHQ \citep{karras2019style} \& MetFaces \citep{metfaces}.} In low and high resolution versions, we center cropped the data with $88/128$ and $800/1024$ factors, respectively, and scaled to the specified resolution.
    \item \textbf{AFHQ-Dog \citep{afhq}.} The images will be resized to the corresponding resolution.
    \item \textbf{Stanford Dogs \citep{stanforddog}.} We crop the images following the official annotations and resize them to the corresponding resolution.
\end{itemize}

\subsubsection{Dataset Splitting.}
This step is used to model training, including classifiers and conditional generators. 
For labeled datasets like CelebA and FaceScrub, we use 80\% data for training, 10\% for validation, and 10\% for testing. The validation set is used to select the best trained models, hyperparameters for training, and hyperparameters for defenses.
For unlabeled datasets such as FFHQ and MetFaces, we provide a script for pseudo-labeling all images by computing the scores. The strategy of PLGMI \citep{plg} for the score calculation is contained in the examples of our toolbox.

\subsubsection{Dataset Synthesis.}
Some methods use synthetic datasets for model training. We provide a script to synthesis datasets by pre-trained GANs according to LOKT \citep{lokt}.

\subsection{Implementation of Attack Models.}
\label{appendix:attack}

By modifying the architecture of the generators and discriminators following PPA \cite{ppa}, we implemented $64\times64$ and $256\times256$ versions for all the custom attack models defined by the official code of implemented algorithms. For those algorithms that use StyleGAN2 \citep{stylegan2}, we provide a wrapper for loading the official model and adapting it to our attack process.

\subsection{Implementation of Classifier Training \& Defense Methods}
\label{Benchmark Details-Details of Classifier Training}

Our toolbox supports the training of a wide range of classifiers, including those used by the official code of most implemented algorithms, as well as those supported by TorchVision.

To ensure consistency across different defense strategies, we provide a unified framework for classifier training. Here we present the general idea of implemented methods.

\begin{enumerate}
    \item \textbf{NO Defense}. The classifiers are trained with the cross-entropy loss function.
    \item \textbf{MID \citep{mid}.} It applies a Gaussian perturbation to the features in front of the classification head to reduce the mutual information of the model inputs and outputs. According to the official code, the method can also be called \textit{VIB}, and we adopt this name in our toolbox. The hyperparameter $\alpha$ controls the factor of the Gaussian perturbation term.
    \item \textbf{BiDO \citep{bido}.} It adds the regular term loss (\textit{COCO} or \textit{HSIC}) function so that the intermediate features decrease the mutual information with the input and increase the mutual information with the output. It has two hyperparameters, $\alpha$ and $\beta$. The former is the factor of the regular term loss between inputs and features, and the latter means that between features and model outputs.
    \item \textbf{LS \citep{ls}.} It adds a label smoothing term with a negative smoothing factor $\alpha$ into the cross-entropy loss.
    \item \textbf{TL \citep{tl}.} This method uses a pre-trained model for fine-tuning. 
    The parameters of the previous layers are frozen and those of the later layers will be fine-tuned. In our implementation, the hyperparameter $\alpha$ is defined as a ratio of the number of frozen layers.
    \item \textbf{TTS \citep{TTS}.} 
    TTS first train classifier without defense. After that, the residual shortcuts in the last stages will be removed. Then, the classifier will be continuously trained. The hyperparameter $\alpha$ means the keep ratio of a residual shortcut, and $n$ denotes the number of residual shortcuts to be removed.
    \item \textbf{Purifier \citep{yang2023purifier}.} Despite the lack of open code and details about $\lambda$ and $k$NN, we reproduce it according to their paper. The $\lambda$ denotes the weight of keeping the first label when training the CVAE. If the L2 distance $<0.0002$ between the output and the nearest $k$ members, we swap the first and second labels. The validation set is used as the reference set it needs.
    \item \textbf{SSD \citep{zhuangstealthy}.} It modifies outputs by solving convex optimization problems. $T$ is used for probability calibration and $\varepsilon$ is the distortion bound. The validation set is used to estimate $\pmb{q}^y$.
\end{enumerate}

In order to fairly compare the various defense methods, all defense methods are trained using the same pre-trained model in our experiments.

\subsection{Details of Attack Process}

The attack process follows a sequential workflow, containing \textbf{latent vectors sampling}, \textbf{labeled latent vectors selection}, \textbf{optimization} and \textbf{final images selection}. Here are the details of the workflow.

\paragraph{Latent Vectors Sampling.}
This step generates random latent vectors and distributes them for each attack target. Most attack methods use a random distribution strategy. Mirror, PPA, and BREP calculate the score of each latent vector corresponding to each label, and for each label,l the vectors with the highest scores are selected. 

\paragraph{Labeled Latent Vectors Selection (Optional).}
The previous step distributes latent vectors for each label, and this step optimizes the latent vectors by calculating the scores of the latent vectors corresponding to the labels and selecting the few vectors with the highest scores. Although currently there are no algorithms that use this step, it can be added into the attack algorithms that use conditional generators, e.g., PLGMI \citep{plg} and LOKT \citep{lokt}. 

\paragraph{Optimization.}
Optimization is the key step for the attack process. It accepts the initialized latent vectors and attack labels as input and outputs the optimized inverted images. We provide several kinds of optimization strategies of each attack method as follows.

\begin{enumerate}
    \item \textbf{Simple White-Box Optimization.} An optimizer for attack algorithms that use the gradient except KEDMI \citep{ked} and VMI \citep{vmi}. It optimize and generates an image for each input vector. Some implement details are displayed in Table \ref{table:white-optimize}.
    \item \textbf{Variance White-Box Optimization.} It optimizes results from a Gaussian distribution of latent vectors corresponding to the target labels, and images are generated by random sampling from the optimized distribution. It is used by KEDMI \citep{ked}. 
    \item \textbf{Miner WhiteBox Optimization.} This optimizer aims to iteratively update the parameters of networks that are utilized to produce high-quality latent vectors, such as Flow models \citep{xu2022poisson}. It is used by VMI \citep{vmi}.
    \item \textbf{Genetic Optimization.} An optimizer using genetic algorithms for optimization. The black-box version of Mirror \citep{mirror} and C2FMI \citep{c2f} use this optimization strategy.
    \item \textbf{BREP Optimization.} A specific optimizer for BREPMI algorithm \citep{brep}. It uses a boundary repelling strategy for gradient simulation.
    \item \textbf{Reinforcement Learning Optimization.} Optimizing the latent vectors via reinforcement learning. Used by RLB attack method \citep{rlb}.
\end{enumerate}

\paragraph{Final Images Selection (Optional).}
This step works by calculating the scores of each image and selecting the part of the image with the highest score as the result of the attack. It is used by PPA \citep{ppa}.

\begin{table}[!ht]
    \setlength{\tabcolsep}{5pt}
    \normalsize
    \centering
    \caption{ Overview of implement of different attack methods that use White-Box Optimization.  }
    \label{table:white-optimize}
    \begin{threeparttable} 
    \resizebox{\linewidth}{!}{
    \begin{tabular}{ccccccc}
        \toprule
        \textbf{Method} & \textbf{Latent Optimizer} & \textbf{Identity Loss} & \textbf{Prior Loss} & \textbf{Image Augment}  \\ \midrule
        GMI & Momentum SGD & CE & Discriminator &  \XSolidBrush \\ 
        KEDMI & Adam & CE & Discriminator & \XSolidBrush   \\ 
        Mirror & Adam & CE& -  &  \CheckmarkBold \\ 
        PPA & Adam & Poincar\'e & -  &  \CheckmarkBold \\ 
        PLGMI & Adam & Max Margin & -  &  \CheckmarkBold \\ 
        LOMMA & Adam & Logit & Feature Distance  & \XSolidBrush  \\ 
        IF-GMI & Adam & Poincar\'e & - & \CheckmarkBold \\ 
        LOKT & Adam & Max Margin & -  &  \CheckmarkBold \\ 
        \bottomrule
    \end{tabular}
    }
    \end{threeparttable}
\end{table}
\section{Evaluation Details}
\label{B}

This section describes the setup and details of the experiments in this paper.

\subsection{Evaluation Settings}
\label{setting}

Following the standard settings in existing works, we conducted experiments at both low and high resolution scenarios. For the low-resolution experiments, we employed classifiers with a resolution of $64\times 64$ as the target models, and a ResNet-$50$ with a resolution of $112\times112$ served as the evaluation model. In the high resolution experiments, we used classifiers with a resolution of $224\times224$ as the target models, with an Inception-v3 model having a resolution of $299\times299$ as the evaluation model. Additionally, the computation resources utilized in our experiments including $16\times$ NVIDIA RTX 4090 and $8\times$ NVIDIA A100.

\subsection{Datasets}
\label{appendix:dataset}

The datasets used in our experiments are categorized into two types: public datasets and private datasets. The private datasets are used to train the target and evaluation models. Specifically, we selected $1000$ identities with the most images from the CelebA dataset and all $530$ identities from the FaceScrub dataset as our private datasets.

The public datasets serve as a priori knowledge for the attacker to train the generator or to extract features of real faces. For the low-resolution experiments, we used FFHQ and the images from the CelebA dataset that are not included in the private dataset as our public datasets. For the high-resolution experiments, FFHQ and MetFaces were chosen as the public datasets. Note that MetFaces is an image dataset of 1336 human faces extracted from the Metropolitan Museum of Art Collection. It has a huge distribution shift with real human faces, which makes model inversion attack algorithms encounter great challenges.

The preprocessing of these datasets is described in \ref{Benchmark Details Data Pro-processing}.

\subsection{Classifiers}

For the attack experiments, we trained multiple classifiers as target models and evaluation models, as detailed in Table \ref{table:classifier-raw}. For the defense experiments, we trained the ResNet-$152$ model on the FaceScrub dataset using various defense methods, as outlined in Table \ref{table:classifier-defense}.

\begin{table}[!ht]
    \setlength{\tabcolsep}{5pt}
    \normalsize
    \centering
    \caption{ Overview of target and evaluation models used in attack experiments. }
    \label{table:classifier-raw}
    \begin{threeparttable} 
    \resizebox{\linewidth}{!}{
    \begin{tabular}{ccccccc}
        \toprule
        \textbf{Dataset} & \textbf{Model} & \textbf{Resolution} & \textbf{Test Acc}  \\ \midrule
        \multirow{5}{*}{CelebA}& VGG-16 & $64\times64$ & 0.882  \\ 
        ~ & ResNet-152 & $64\times64$ & 0.937  \\ 
        ~ & ResNet-50 & $112\times112$ & 0.958  \\ 
        ~ & ResNet-152 & $224\times224$ & 0.900  \\ 
        ~ & Inception-v3 & $299\times299$ & 0.921  \\ \midrule
        \multirow{6}{*}{FaceScrub} & VGG-16 & $64\times64$ & 0.878  \\ 
        ~ & ResNet-152 & $64\times64$ & 0.982  \\ 
        ~ & ResNet-50 & $112\times112$ & 0.993  \\ 
        ~ & ResNet-152 & $224\times224$ & 0.922  \\ 
        ~ & ResNeSt-101 & $224\times224$ & 0.932  \\ 
        ~ & Inception-v3 & $299\times299$ & 0.944  \\ 
        \bottomrule
    \end{tabular}
    }
    \end{threeparttable}
\end{table}
\begin{table}[!ht]
    \setlength{\tabcolsep}{5pt}
    \normalsize
    \centering
    \caption{ Overview of IR-$152$ trained with FaceScrub dataset in different defense methods. The definition of hyperparameters are described in \ref{Benchmark Details-Details of Classifier Training}.  }
    \label{table:classifier-defense}
    \begin{threeparttable} 
    \resizebox{\linewidth}{!}{
    \begin{tabular}{ccccccc}
        \toprule
        \textbf{Defense Method} & \textbf{Hyperparameters} & \textbf{Test Acc}  \\ \midrule
        No Defense & - &$0.982$  \\ 
        MID & $\alpha=0.01$ & $0.982$  \\ 
        BiDO & $\alpha=0.01,\beta= 0.1$ & $0.940$  \\ 
        LS & $\alpha=-0.05$ & $0.980$  \\ 
        TL & $\alpha=0.5$ & $0.953$  \\ 
        TTS & $\alpha=0,n=3$ & $0.973$ \\
        Purifier & $\lambda=0.01,k=1$ & $0.946$ \\
        SSD & $T=0.05,\varepsilon=1$ & $0.950$ \\
        \bottomrule
    \end{tabular}
    }
    \end{threeparttable}
\end{table}

\subsection{Evaluation Metrics}

The definitions of the evaluation metrics are detailed in Section \ref{Section:benchmark: Details of Evaluation}. Here, we present the specific details of the metrics used in the experiments.

For the Classification Accuracy and Feature Distance metrics, we evaluate the attack results using another classifier pre-trained on the same dataset as the target model: ResNet-50 for low-resolution settings and Inception-v3 for high-resolution settings, denoted as \textbf{Acc} and $\delta_{eval}$. Additionally, an Inception-v1 model pre-trained on a large face dataset, VGGFace2, is used to calculate the feature distance, measuring the realism of the inverted images, denoted as $\delta_{face}$.

For FID, Precision-Recall, and Density-Coverage scores, we follow the experimental setup of existing papers. We use Inception-v3, pre-trained on ImageNet, to extract the features of images and participate in the score calculation.

\section{More Evaluation}
\label{more_evaluation}
\subsection{Sample Diversity}
\label{prdc}

Following the settings of Section \ref{section:Evaluation on Different Attack Methods} and \ref{section:Evaluation on Higher Resolution}, we computed the Precision-Recall \citep{pr} and Density-Coverage \citep{pr} to evaluate the intra-class diversity for each attack method. The results are presented in Table \ref{table:white-prdc-ffhq64_facescrub64_ir152}, \ref{table:black-prdc-ffhq64_facescrub64_ir152}, \ref{table:white-prdc-ffhq256_facescrub64_res152} and \ref{table:black-prdc-ffhq256_facescrub64_res152}.  It tends to be that the attacks with stronger GAN priors get higher scores, such as Mirror, C2FMI and PPA.

\begin{table}[!ht]
    \setlength{\tabcolsep}{5pt}
    \normalsize
    \centering
    \caption{Comparison between white-box MI attacks on low resolution scenario.}
    \label{table:white-prdc-ffhq64_facescrub64_ir152}
    \begin{threeparttable} 
    \resizebox{\linewidth}{!}{
    \begin{tabular}{ccccccc}
        \toprule
\textbf{Method} & $\uparrow\textbf{Precision}$ & $\uparrow\textbf{Recall}$ & $\uparrow\textbf{Density}$ & $\uparrow\textbf{Coverage}$\\ \midrule
GMI & $0.025\pm0.079$ & $0.797\pm0.200$ & $0.008\pm0.018$ & $0.019\pm0.037$\\
KEDMI & $0.065\pm0.159$ & $0.055\pm0.158$ & $0.017\pm0.051$ & $0.025\pm0.062$\\
Mirror(white) & $\mathbf{0.205}\pm0.232$ & $0.444\pm0.304$ & $\mathbf{0.067}\pm0.095$ & $\mathbf{0.133}\pm0.148$\\
PPA & $0.149\pm0.199$ & $0.499\pm0.273$ & $0.043\pm0.073$ & $0.089\pm0.122$\\
PLGMI & $0.048\pm0.116$ & $0.302\pm0.283$ & $0.014\pm0.038$ & $0.030\pm0.064$\\
LOMMA+GMI & $0.062\pm0.123$ & $\mathbf{0.828}\pm0.187$ & $0.018\pm0.043$ & $0.040\pm0.080$\\
LOMMA+KEDMI & $0.061\pm0.187$ & $0.000\pm0.008$ & $0.019\pm0.052$ & $0.023\pm0.052$\\
IF-GMI & $0.129\pm0.191$ & $0.585\pm0.279$ & $0.039\pm0.068$ & $0.081\pm 0.115$\\
\bottomrule
    \end{tabular}
    }
    \end{threeparttable}
\end{table}
\begin{table}[!ht]
    \setlength{\tabcolsep}{5pt}
    \normalsize
    \centering
    \caption{Comparison between black-box MI attacks on low resolution scenario.}
    \label{table:black-prdc-ffhq64_facescrub64_ir152}
    \begin{threeparttable} 
    \resizebox{\linewidth}{!}{
    \begin{tabular}{ccccccc}
        \toprule
\textbf{Method} & $\uparrow\textbf{Precision}$ & $\uparrow\textbf{Recall}$ & $\uparrow\textbf{Density}$ & $\uparrow\textbf{Coverage}$\\ \midrule
BREP & $0.048\pm0.131$ & $0.249\pm0.309$ & $0.013\pm0.030$ & $0.023\pm0.051$\\
Mirror(black) & $0.085\pm0.147$ & $\mathbf{0.489}\pm0.294$ & $\mathbf{0.262}\pm0.043$ & $\mathbf{0.059}\pm0.083$\\
C2F & $\mathbf{0.118}\pm0.234$ & $0.029\pm0.125$ & $0.037\pm0.078$ & $0.053\pm0.089$\\
LOKT & $0.051\pm0.129$ & $0.232\pm0.292$ & $0.013\pm0.032$ & $0.027\pm0.063$\\
\bottomrule
    \end{tabular}
    }
    \end{threeparttable}
\end{table}

\begin{table}[!ht]
    \setlength{\tabcolsep}{5pt}
    \normalsize
    \centering
    \caption{Comparison between white-box MI attacks on high resolution scenario.}
    \label{table:white-prdc-ffhq256_facescrub64_res152}
    \begin{threeparttable} 
    \resizebox{\linewidth}{!}{
    \begin{tabular}{ccccccc}
        \toprule
\textbf{Method} & $\uparrow\textbf{Precision}$ & $\uparrow\textbf{Recall}$ & $\uparrow\textbf{Density}$ & $\uparrow\textbf{Coverage}$\\ \midrule
GMI & $0.033\pm0.086$ & $\mathbf{0.758}\pm0.248$ & $0.005\pm0.011$ & $0.013\pm0.028$\\
KEDMI & $0.029\pm0.116$ & $0.055\pm0.170$ & $0.006\pm0.016$ & $0.010\pm0.027$\\
Mirror(white) & $0.217\pm0.236$ & $0.350\pm0.292$ & $0.048\pm0.072$ & $0.092\pm0.099$\\
PPA & $\mathbf{0.259}\pm0.243$ & $0.322\pm0.266$ & $\mathbf{0.060}\pm0.075$ & $\mathbf{0.112}\pm0.102$\\
PLGMI & $0.019\pm0.099$ & $0.002\pm0.025$ & $0.005\pm0.015$ & $0.007\pm0.018$\\
LOMMA+GMI & $0.023\pm0.074$ & $0.514\pm0.333$ & $0.006\pm0.013$ & $0.013\pm0.028$\\
LOMMA+KEDMI & $0.033\pm0.130$ & $0.003\pm0.041$ & $0.008\pm0.026$ & $0.011\pm0.024$\\
IF-GMI & $0.154\pm0.200$ & $0.339\pm0.287$ & $0.035\pm0.052$ & $0.068\pm0.074$ \\
\bottomrule
    \end{tabular}
    }
    \end{threeparttable}
\end{table}
\begin{table}[!ht]
    \setlength{\tabcolsep}{5pt}
    \normalsize
    \centering
    \caption{Comparison between black-box MI attacks on high resolution scenario.}
    \label{table:black-prdc-ffhq256_facescrub64_res152}
    \begin{threeparttable} 
    \resizebox{\linewidth}{!}{
    \begin{tabular}{ccccccc}
        \toprule
\textbf{Method} & $\uparrow\textbf{Precision}$ & $\uparrow\textbf{Recall}$ & $\uparrow\textbf{Density}$ & $\uparrow\textbf{Coverage}$\\ \midrule
BREP & $0.041\pm0.121$ & $\mathbf{0.160}\pm0.286$ & $0.008\pm0.024$ & $0.014\pm0.029$\\
Mirror(black) & $0.011\pm0.055$ & $0.115\pm0.201$ & $0.004\pm0.010$ & $0.008\pm0.024$\\
C2F & $\mathbf{0.119}\pm0.227$ & $0.026\pm0.115$ & $\mathbf{0.024}\pm0.048$ & $\mathbf{0.036}\pm0.063$\\
LOKT & $0.014\pm0.083$ & $0.023\pm0.125$ & $0.004\pm0.013$ & $0.006\pm0.016$\\
\bottomrule
    \end{tabular}
    }
    \end{threeparttable}
\end{table}

\subsection{Evaluation on Different Target Classifiers}

In addition to attacking IR-152 and ResNet-152 in Section \ref{section:Evaluation on Different Attack Methods} and \ref{section:Evaluation on Higher Resolution}, we extend our experiments on more different target classifiers. We evaluate attacks on VGG-$16$ \citep{vgg} and ResNeSt-$101$ \citep{zhang2022resnest} in low- and high-resolution settings, respectively. The results are presented in Table \ref{table:vgg16-attack} and \ref{table:resnest-attack}. 

\begin{table}[!ht]
    \setlength{\tabcolsep}{5pt}
    \normalsize
    \centering
    \caption{Comparison between white-box MI attacks against VGG-$16$ on low resolution scenario.}
    \label{table:vgg16-attack}
    \begin{threeparttable} 
    \resizebox{\linewidth}{!}{
    \begin{tabular}{ccccccc}
        \toprule
\textbf{Method} & $\uparrow{\mathbf{Acc}@1}$ & $\uparrow{\mathbf{Acc}@5}$ & $\downarrow\mathbf{\delta}_{eval}$ & $\downarrow\mathbf{\delta}_{face}$ & $\downarrow\textbf{FID}$\\ \midrule
GMI & $0.013\pm0.003$ & $0.046\pm0.018$ & $2565.303\pm290.350$ & $1.352\pm0.142$ & $62.205$\\
KEDMI & $0.074\pm0.008$ & $0.190\pm0.013$ & $2553.729\pm412.648$ & $1.147\pm0.254$ & $91.953$\\
Mirror(white) & $0.061\pm0.007$ & $0.165\pm0.006$ & $2358.875\pm347.703$ & $1.111\pm0.174$ & $37.605$\\
PPA & $0.263\pm0.019$ & $0.461\pm0.023$ & $2018.148\pm377.491$ & $0.874\pm0.160$ & $\mathbf{33.226}$\\
PLGMI & $\mathbf{0.465}\pm0.019$ & $\mathbf{0.683}\pm0.008$ & $\mathbf{1914.942}\pm409.569$ & $\mathbf{0.762}\pm0.174$ & $81.093$\\
LOMMA+GMI & $0.091\pm0.026$ & $0.216\pm0.047$ & $2503.465\pm288.728$ & $1.060\pm0.153$ & $60.650$\\
LOMMA+KEDMI & $0.233\pm0.009$ & $0.418\pm0.011$ & $2258.070\pm480.906$ & $0.912\pm0.205$ & $66.410$\\
IF-GMI & $0.208\pm0.010$ & $0.391\pm0.021$ & $2102.656\pm369.571$ & $0.928\pm0.172$ & $35.816$ \\
\bottomrule
    \end{tabular}
    }
    \end{threeparttable}
\end{table}

\begin{table}[!ht]
    \setlength{\tabcolsep}{5pt}
    \normalsize
    \centering
    \caption{Comparison between white-box MI attacks against ResNeSt-$101$ on high resolution scenario.}
    \label{table:resnest-attack}
    \begin{threeparttable} 
    \resizebox{\linewidth}{!}{
    \begin{tabular}{ccccccc}
        \toprule
\textbf{Method} & $\uparrow{\mathbf{Acc}@1}$ & $\uparrow{\mathbf{Acc}@5}$ & $\downarrow\mathbf{\delta}_{eval}$ & $\downarrow\mathbf{\delta}_{face}$ & $\downarrow\textbf{FID}$\\ \midrule
GMI & $0.069\pm0.011$ & $0.191\pm0.036$ & $135.290\pm22.961$ & $1.339\pm0.135$ & $124.880$\\
KEDMI & $0.153\pm0.013$ & $0.353\pm0.012$ & $143.155\pm32.520$ & $1.258\pm0.245$ & $140.533$\\
Mirror(white) & $0.380\pm0.027$ & $0.684\pm0.021$ & $193.275\pm29.316$ & $1.032\pm0.161$ & $58.437$\\
PPA & $0.904\pm0.008$ & $0.984\pm0.002$ & $159.986\pm27.495$ & $0.781\pm0.157$ & $\mathbf{44.966}$\\
PLGMI & $\mathbf{0.931}\pm0.006$ & $\mathbf{0.988}\pm0.003$ & $147.914\pm40.333$ & $\mathbf{0.753}\pm0.177$ & $92.755$\\
LOMMA+GMI & $0.577\pm0.134$ & $0.770\pm0.123$ & $\mathbf{131.040}\pm28.470$ & $1.042\pm0.165$ & $133.604$\\
LOMMA+KEDMI & $0.373\pm0.008$ & $0.615\pm0.007$ & $148.923\pm42.489$ & $1.129\pm0.285$ & $139.433$\\
IF-GMI & $0.736\pm 0.013$ & $0.920\pm 0.011$ & $236.910\pm 49.451$ & $0.647\pm 0.133$ & $45.759$ \\
\bottomrule
    \end{tabular}
    }
    \end{threeparttable}
\end{table}
\begin{table}[!ht]
    \setlength{\tabcolsep}{5pt}
    \normalsize
    \centering
    \caption{Comparison between white-box MI attacks against MaxViT on high resolution scenario.}
    \label{table:vit-attack}
    \begin{threeparttable} 
    \resizebox{\linewidth}{!}{
    \begin{tabular}{ccccccc}
        \toprule
\textbf{Method} & $\uparrow{\mathbf{Acc}@1}$ & $\uparrow{\mathbf{Acc}@5}$ & $\downarrow\mathbf{\delta}_{eval}$ & $\downarrow\mathbf{\delta}_{face}$ & $\downarrow\textbf{FID}$\\ \midrule
GMI & $0.018 \pm 0.007$ & $0.080 \pm 0.021$ & $260.084 \pm 71.029$ & $1.406 \pm 0.131$ & $154.447$ \\
KEDMI & $0.112 \pm 0.007$ & $0.270 \pm 0.028$ & $261.827 \pm 73.975$ & $1.117 \pm 0.224$ & $148.083$ \\
Mirror & $0.146 \pm 0.034$ & $0.346 \pm 0.066$ & $286.339 \pm 47.047$ & $1.034 \pm 0.158$ & $80.136$ \\
PPA & $0.522 \pm 0.020$ & $0.758 \pm 0.010$ & $237.410 \pm 41.175$ & $0.776 \pm 0.118$ & $66.023$ \\
PLGMI & $0.322 \pm 0.035$ & $0.574 \pm 0.042$ & $261.860 \pm 55.469$ & $0.772 \pm 0.137$ & $153.054$ \\
LOMMA+GMI & $0.374 \pm 0.077$ & $0.620 \pm 0.058$ & $244.566 \pm 48.069$ & $0.920 \pm 0.148$ & $138.875$ \\
LOMMA+KEDMI & $0.294 \pm 0.014$ & $0.552 \pm 0.022$ & $260.002 \pm 71.687$ & $0.910 \pm 0.217$ & $150.214$ \\
IF-GMI & $0.408 \pm 0.012$ & $0.669 \pm 0.020$ & $230.251 \pm 42.734$ & $0.801 \pm 0.139$ & $45.625$ \\
\bottomrule
    \end{tabular}
    }
    \end{threeparttable}
\end{table}

Besides the aforementioned CNN-based classifiers, we further analyze MI attacks on ViT-based models. Table \ref{table:vit-attack} presents further experiments on the MaxViT \citep{tu2022maxvit} under the high resolution (224×224) scenario. Other experimental settings are continuous with the main paper, with FFHQ as the public dataset and FaceScrub as the private dataset. The test accuracy of the target MaxViT is 94.61\%.

\subsection{Evaluation on More Combination of Datasets}
\label{combination}
Evaluations in the Section \ref{experiment} are conducted under the same dataset combination of FFHQ as the public dataset and FaceScrub as the private dataset. Therefore, we design more combination of datasets in this part to further assess the transferability of different attacks. The results are listed in Table \ref{table:ffhq-celeba} and \ref{table:metfaces-celeba}. The visual results are shown in Figure \ref{visualffhq} and \ref{visualmetfaces}.

Except for the typical face classification task, we have conducted more experiments on the dog breed classification task under the high resolution (224×224) scenario, which includes two non-facial datasets, Stanford Dogs \citep{standforddog} and Animal Faces-HQ Dog (AFHQ) \citep{afhq}. The public dataset is AFHQ while the private dataset is Stanford Dogs. The target model is ResNet-152 of 77.45\% test accuracy, which follows the same setting in PPA. Table \ref{table:dog} lists the evaluation results.

\begin{table}[!ht]
    \setlength{\tabcolsep}{5pt}
    \normalsize
    \centering
    \caption{Comparison between white-box MI attacks with FFHQ prior against ResNet-$152$ pre-trained on CelebA on high resolution scenario.}
    \label{table:ffhq-celeba}
    \begin{threeparttable} 
    \resizebox{\linewidth}{!}{
    \begin{tabular}{ccccccc}
        \toprule
\textbf{Method} & $\uparrow{\mathbf{Acc}@1}$ & $\uparrow{\mathbf{Acc}@5}$ & $\downarrow\mathbf{\delta}_{eval}$ & $\downarrow\mathbf{\delta}_{face}$ & $\downarrow\textbf{FID}$\\ \midrule
GMI & $0.050\pm0.008$ & $0.171\pm0.031$ & $216.614\pm36.221$ & $1.248\pm0.138$ & $108.217$\\
KEDMI & $0.174\pm0.013$ & $0.391\pm0.011$ & $247.112\pm55.473$ & $1.113\pm0.221$ & $119.760$\\
Mirror(white) & $0.367\pm0.026$ & $0.661\pm0.024$ & $286.668\pm47.261$ & $0.973\pm0.166$ & $63.261$\\
PPA & $0.936\pm0.008$ & $0.987\pm0.002$ & $233.474\pm50.366$ & $\mathbf{0.711}\pm0.148$ & $\mathbf{46.339}$\\
PLGMI & $\mathbf{0.953}\pm0.007$ & $\mathbf{0.992}\pm0.002$ & $261.210\pm58.636$ & $0.726\pm0.167$ & $151.119$\\
LOMMA+GMI & $0.664\pm0.121$ & $0.815\pm0.100$ & $\mathbf{207.854}\pm39.254$ & $0.938\pm0.162$ & $109.383$\\
LOMMA+KEDMI & $0.222\pm0.005$ & $0.411\pm0.007$ & $229.407\pm65.371$ & $1.178\pm0.346$ & $145.272$\\
IF-GMI & $0.986\pm 0.003$ & $0.999\pm 0.001$ & $222.919\pm52.073$ & $0.614\pm 0.140$ & $37.408$ \\
\bottomrule
    \end{tabular}
    }
    \end{threeparttable}
\end{table}
\begin{table}[!ht]
    \setlength{\tabcolsep}{5pt}
    \normalsize
    \centering
    \caption{Comparison between white-box MI attacks with Metfaces prior against ResNet-$152$ pre-trained on CelebA on high resolution scenario.}
    \label{table:metfaces-celeba}
    \begin{threeparttable} 
    \resizebox{\linewidth}{!}{
    \begin{tabular}{ccccccc}
        \toprule
\textbf{Method} & $\uparrow{\mathbf{Acc}@1}$ & $\uparrow{\mathbf{Acc}@5}$ & $\downarrow\mathbf{\delta}_{eval}$ & $\downarrow\mathbf{\delta}_{face}$ & $\downarrow\textbf{FID}$\\ \midrule
GMI & $0.008\pm0.003$ & $0.046\pm0.008$ & $\mathbf{209.264}\pm45.093$ & $1.392\pm0.149$ & $217.151$\\
KEDMI & $0.002\pm0.001$ & $0.011\pm0.002$ & $250.805\pm62.654$ & $1.561\pm0.232$ & $276.504$\\
Mirror(white) & $0.100\pm0.007$ & $0.265\pm0.009$ & $357.719\pm52.080$ & $1.261\pm0.194$ & $78.541$\\
PPA & $\mathbf{0.463}\pm0.020$ & $\mathbf{0.726}\pm0.020$ & $305.953\pm57.145$ & $\mathbf{1.074}\pm0.203$ & $\mathbf{72.372}$\\
PLGMI & $0.126\pm0.003$ & $0.274\pm0.005$ & $220.139\pm41.739$ & $1.126\pm0.218$ & $393.518$\\
LOMMA+GMI & $0.061\pm0.019$ & $0.140\pm0.032$ & $214.122\pm54.770$ & $1.370\pm0.228$ & $245.013$\\
LOMMA+KEDMI & $0.006\pm0.001$ & $0.013\pm0.001$ & $245.896\pm63.101$ & $1.630\pm0.253$ & $320.662$\\
IF-GMI & $0.934\pm 0.010$ & $0.988\pm 0.003$ & $235.986\pm 46.216$ & $0.768\pm0.162$ & $73.375$ \\
\bottomrule
    \end{tabular}
    }
    \end{threeparttable}
\end{table}
\begin{table}[!ht]
    \setlength{\tabcolsep}{5pt}
    \normalsize
    \centering
    \caption{Comparison between white-box MI attacks with AFHQ prior against ResNet-152 pre-trained on Stanford Dogs on high resolution scenario.}
    \label{table:dog}
    \begin{threeparttable} 
    \resizebox{.9\linewidth}{!}{
    \begin{tabular}{cccccc}
        \toprule
\textbf{Method} & $\uparrow{\mathbf{Acc}@1}$ & $\uparrow{\mathbf{Acc}@5}$ & $\downarrow\mathbf{\delta}_{eval}$ & $\downarrow\textbf{FID}$\\ \midrule
GMI & $0.068 \pm 0.031$ & $0.226 \pm 0.026$ & $88.447 \pm 15.990$ & $210.543$ \\
KEDMI & $0.606 \pm 0.027$ & $0.830 \pm 0.032$ & $66.521 \pm 16.994$ & $134.513$ \\
Mirror & $0.656 \pm 0.058$ & $0.848 \pm 0.017$ & $142.580 \pm 49.569$ & $77.485$ \\
PPA & $0.906 \pm 0.026$ & $0.990 \pm 0.006$ & $121.571 \pm 45.929$ & $58.479$ \\
PLGMI & $0.216 \pm 0.016$ & $0.504 \pm 0.022$ & $86.629 \pm 20.109$ & $238.115$ \\
LOMMA+GMI & $0.302 \pm 0.103$ & $0.486 \pm 0.126$ & $84.761 \pm 24.458$ & $198.523$ \\
LOMMA+KEDMI & $0.838 \pm 0.007$ & $0.968 \pm 0.007$ & $\textbf{58.225} \pm 22.527$ & $97.301$ \\
IF-GMI & $\textbf{0.947} \pm 0.008$ & $\textbf{0.993} \pm 0.003$ & $147.845 \pm 66.393$ & $\textbf{48.972}$ \\
\bottomrule
    \end{tabular}
    }
    \end{threeparttable}
\end{table}

\subsection{Evaluation on Different Loss Functions}

In recent years, various attack algorithms have attempted to mitigate the effects of gradient vanishing by employing different loss functions. In this part, we investigate the impact of identity loss functions on the success rate of model inversion attacks. Specifically, we adopt PPA with FFHQ prior to attack a ResNet-152 classifier pre-trained on FaceScrub. The comparison is presented in Table \ref{table:diff loss}. Our findings indicate that the Poincar\'e loss function yields the highest performance without model augmentation, whereas the Logit loss function achieves the best results with model augmentation, serving as the optimal loss choices for future studies. 

\begin{table}[!ht]
    \setlength{\tabcolsep}{5pt}
    \normalsize
    \centering
    \caption{Comparision of different identity loss. "$+$" denotes the target model is used as the teacher model, and three students models are distilled using the public dataset and jointly involved in the loss calculation. It is called model augmentation in the paper of \cite{lomma}. Logit loss here is implemented via pytorch's NLLLoss.}
    \label{table:diff loss}
    \begin{threeparttable} 
    \resizebox{\linewidth}{!}{
    \begin{tabular}{ccccccc}
        \toprule
\textbf{Loss Function} & $\uparrow{\mathbf{Acc}@1}$ & $\uparrow{\mathbf{Acc}@5}$ & $\downarrow\mathbf{\delta}_{eval}$ & $\downarrow\mathbf{\delta}_{face}$ & $\downarrow\textbf{FID}$\\ \midrule
CE & $0.769\pm0.032$ & $0.942\pm0.012$ & $172.305\pm26.753$ & $0.901\pm0.140$ & $53.880$\\
Poincar\'e & $0.913\pm0.022$ & $0.986\pm0.004$ & $167.532\pm28.944$ & $0.774\pm0.143$ & $46.246$\\
Max Margin & $0.812\pm0.020$ & $0.951\pm0.008$ & $169.730\pm27.705$ & $0.871\pm0.150$ & $51.146$\\
Logit & $0.886\pm0.023$ & $0.978\pm0.013$ & $170.867\pm31.415$ & $0.806\pm0.148$ & $45.731$\\
CE$^+$ & $\mathbf{0.946}\pm0.011$ & $0.992\pm0.002$ & $165.461\pm29.055$ & $0.785\pm0.147$ & $48.564$\\
Poincar\'e$^+$ & $0.901\pm0.006$ & $0.984\pm0.004$ & $\mathbf{153.921}\pm24.542$ & $0.812\pm0.158$ & $45.114$\\
Max Margin$^+$ & $0.918\pm0.017$ & $0.985\pm0.003$ & $165.420\pm27.607$ & $0.815\pm0.148$ & $49.292$\\
Logit$^+$ & $0.945\pm0.007$ & $\mathbf{0.993}\pm0.003$ & $177.745\pm34.030$ & $\mathbf{0.764}\pm0.159$ & $\mathbf{44.166}$\\
\bottomrule
    \end{tabular}
    }
    \end{threeparttable}
\end{table}

\subsection{Evaluation on Time Overhead}
\label{appendix: time}

We conduct an extra evaluation on the time overhead among all the MI attacks. The results are recorded in the Table \ref{tab:time cost}. The time complexities of most algorithms fall within the same order of magnitude, which is around 100 seconds per class. Notably, the time cost of VMI and RLBMI is extraordinarily higher than other attacks, indicating that both methods are hard to evaluate under the identical scenarios in the main evaluation and relatively unpractical in real-world settings. Therefore, we put the relevant experiments on the next section (Supp.\ref{v-r-p}).

\begin{table}[!ht]
    \setlength{\tabcolsep}{5pt}
    \normalsize
    \centering
    \caption{The time overhead of different attacks when attacking a single class.}
    \label{tab:time cost}
    \begin{threeparttable} 
    \resizebox{.9\linewidth}{!}{
    \begin{tabular}{cccc}
        \toprule
\textbf{Attacks} & $\downarrow\textbf{Time Overhead(seconds)}$   \\ \midrule
\textbf{GMI} & 78 \\
\textbf{KEDMI} & 73 \\
\textbf{VMI} & 5910 \\
\textbf{Mirror(white-box)} & \textbf{43} \\
\textbf{PPA} & 124 \\
\textbf{PLGMI} & 59 \\
\textbf{LOMMA+GMI} & 133 \\
\textbf{LOMMA+KEDMA} & 127 \\
\textbf{IF-GMI} & 83 \\ \midrule
\textbf{Mirror(black-box)} & 161 \\
\textbf{C2FMI} & 83 \\
\textbf{RLBMI} & 10633 \\
\textbf{BREPMI} & 298 \\
\textbf{LOKT} & 219 \\
\bottomrule
    \end{tabular}
}
    \end{threeparttable}
\end{table}

\subsection{More evaluation on VMI, RLBMI and PPA}
\label{v-r-p}

Considering the high computational overhead of RLBMI and VMI, we only experiment at a low resolution settings for $100$ classes. The public and private datasets are splitted part of CelebA dataset. The results are shown in Table \ref{table:rlb vmi}.

\begin{table}[!ht]
    \setlength{\tabcolsep}{5pt}
    \normalsize
    \centering
    \caption{Experimental results of VMI and RLBMI.}
    \label{table:rlb vmi}
    \begin{threeparttable} 
    \resizebox{\linewidth}{!}{
    \begin{tabular}{ccccccc}
        \toprule
\textbf{Method} & $\uparrow{\mathbf{Acc}@1}$ & $\uparrow{\mathbf{Acc}@5}$ & $\downarrow\mathbf{\delta}_{eval}$ & $\downarrow\mathbf{\delta}_{face}$\\ \midrule
VMI & $0.168\pm0.018$ & $0.273\pm0.019$ & $1822.122\pm398.948$ & $1.260\pm0.397$\\
RLBMI & $0.780\pm0.040$ & $0.920\pm0.075$ & $1173.134\pm250.088$ & $0.699\pm0.049$\\
\bottomrule
    \end{tabular}
    }
    \end{threeparttable}
\end{table}

We also explored the effect of PPA for different number of latent vectors for optimization and the number of iterations. It is presented in Table \ref{table: ppa extend}. Note that PPA select top-$5$ optimized latent vectors as attack results.

\begin{table}[!ht]
    \setlength{\tabcolsep}{5pt}
    \normalsize
    \centering
    \caption{Experiment result of PPA for different number of latent vectors to optimize and iterations.}
    \label{table: ppa extend}
    \begin{threeparttable} 
    \resizebox{\linewidth}{!}{
    \begin{tabular}{ccccccc}
        \toprule
\textbf{Number of latents} & \textbf{Iterations} & $\uparrow{\mathbf{Acc}@1}$ & $\uparrow{\mathbf{Acc}@5}$ & $\downarrow\mathbf{\delta}_{eval}$ & $\downarrow\mathbf{\delta}_{face}$ & $\downarrow\textbf{FID}$\\ \midrule
20 & 50 & $0.433\pm0.072$ & $0.651\pm0.076$ & $1888.342\pm478.644$ & $0.898\pm0.239$ & $40.138$\\
50 & 50 & $0.496\pm0.067$ & $0.698\pm0.048$ & $1804.830\pm471.815$ & $0.847\pm0.224$ & $\mathbf{38.765}$\\
20 & 600 & $\mathbf{0.844}\pm0.042$ & $\mathbf{0.924}\pm0.026$ & $\mathbf{1391.261}\pm396.732$ & $\mathbf{0.658}\pm0.194$ & $46.246$\\
\bottomrule
    \end{tabular}
    }
    \end{threeparttable}
\end{table}

\subsection{More Visualization}
\label{appendix: visualization}

\begin{figure*}[tbp]
\centerline{\includegraphics[width=\textwidth]{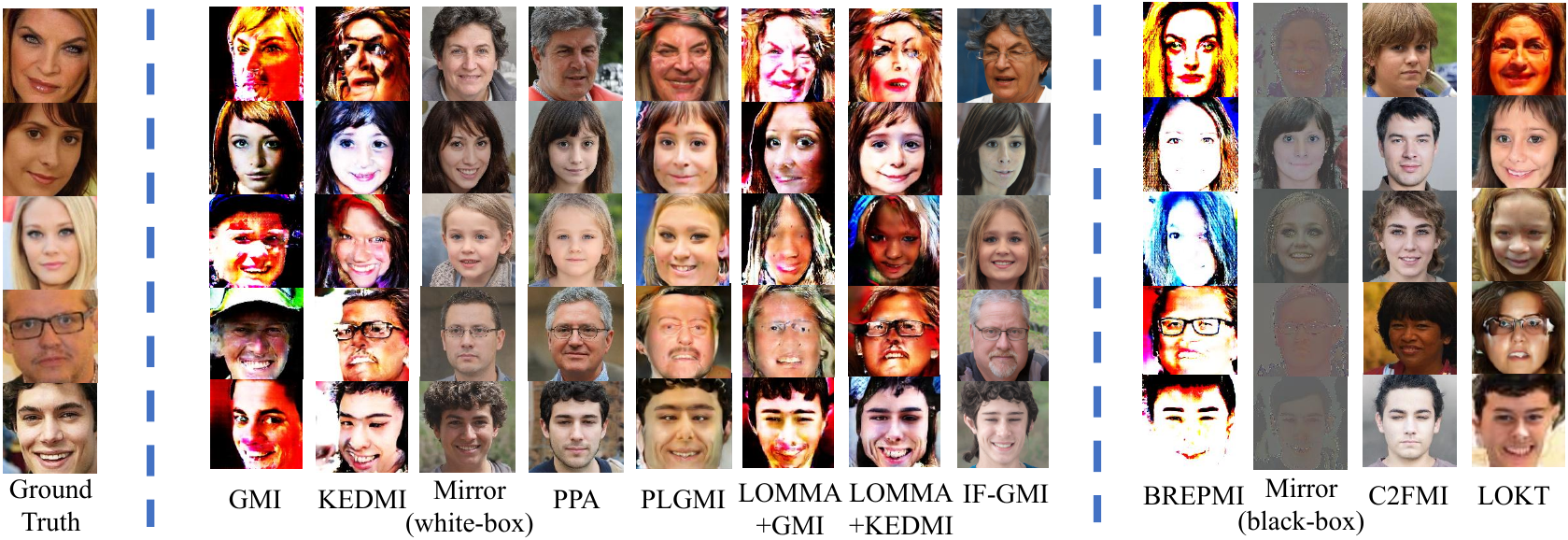}}
\caption{Visual comparison between different MI attacks on higher resolution scenario with FFHQ prior targeting FaceScrub.}
\label{visual-high}
\end{figure*}

\begin{figure*}[!ht]
\centerline{\includegraphics[width=\textwidth]{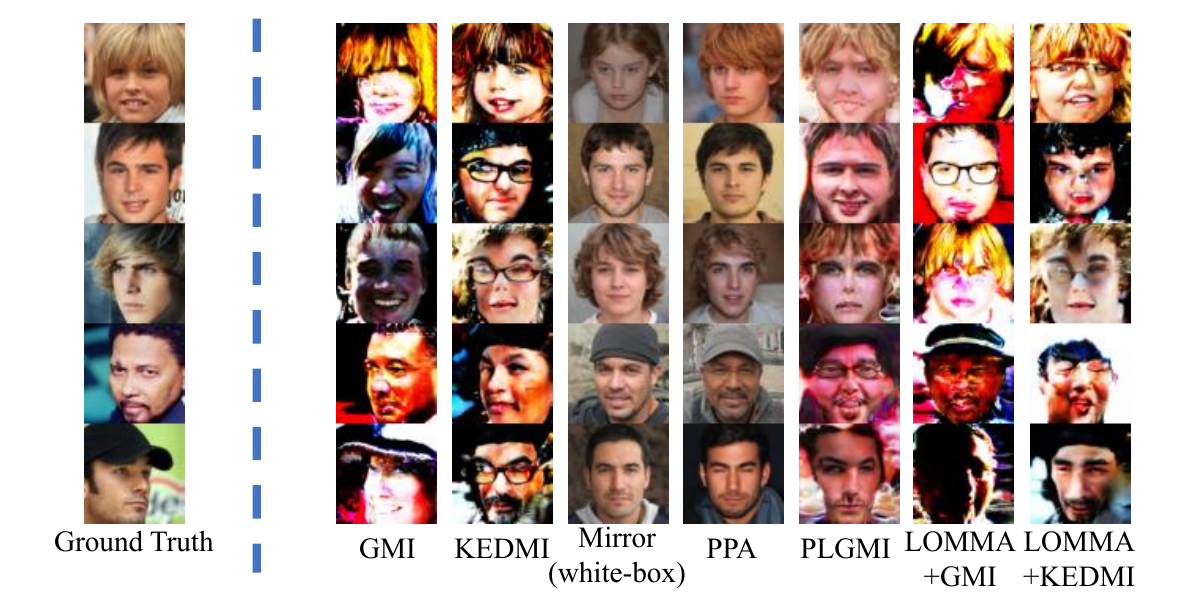}}
\caption{Visual comparison between different MI attacks with FFHQ prior targeting CelebA.}
\label{visualffhq}
\end{figure*}

\begin{figure*}[!ht]
\centerline{\includegraphics[width=\textwidth]{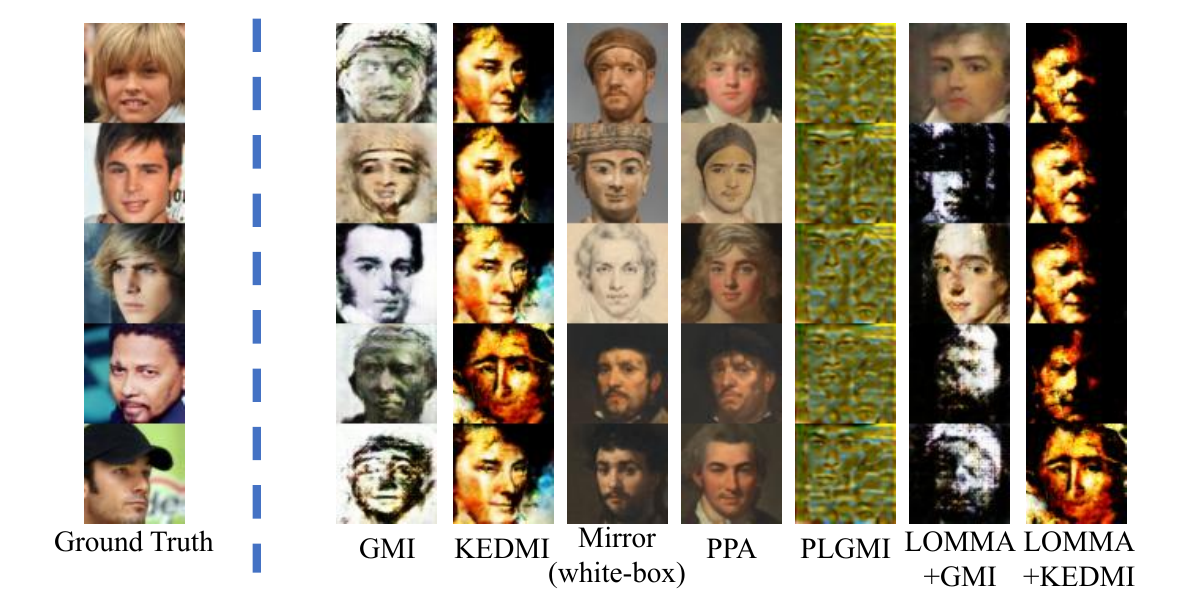}}
\caption{Visual comparison between different MI attacks with MetFaces prior targeting CelebA.}
\label{visualmetfaces}
\end{figure*}

We visualize more results in Figure \ref{visual-high}, \ref{visualffhq} and \ref{visualmetfaces}. The target model is ResNet-$152$ trained on CelebA dataset in high resolution setting.

\section{Special Experiments}
\label{special_exp}

\subsection{Validation on LS Defense Method}
\label{appendix: ls}

To validate the correctness of our implemented LS defense, we follow the settings in the official paper \citep{ls} and evaluate it in more settings, shown in Table \ref{table:ls defense}. To better evaluate the effectiveness of LS defense algorithms, we make the undefended classifiers slightly less accurate than the defense-imposed classifiers through an early-stop strategy when training. 

In setting A, the public and private datasets are different part of CelebA dataset in low resolution scenario with the GMI as the attack method.
In setting B, the public dataset is FFHQ and the private dataset is FaceScrub with the PPA as the attack method.
In these two settings, the predictive power is much lower than that describes in Section \ref{section:Evaluation on Defense Strategies}.  
In this case, the LS defense is very effective, making the success rate of the attack drop dramatically.

\begin{table}[!ht]
    \setlength{\tabcolsep}{5pt}
    \normalsize
    \centering
    \caption{Evaluation on LS defense method on different settings.}
    \label{table:ls defense}
    \begin{threeparttable} 
    \resizebox{\linewidth}{!}{
    \begin{tabular}{ccccccc}
        \toprule
\textbf{Setting} & \textbf{Defense} & \textbf{Test Acc} & $\uparrow{\mathbf{Acc}@1}$ & $\uparrow{\mathbf{Acc}@5}$ & $\downarrow\mathbf{\delta}_{eval}$ & $\downarrow\mathbf{\delta}_{face}$\\ \midrule
\multirow{2}{*}{A} & - & $0.832$ & $0.068\pm0.054$ & $0.180\pm0.126$ & $1949.072\pm184.779$ & $1.281\pm0.129$\\
~ & LS & $0.851$ & $0.005\pm0.003$ & $0.021\pm0.014$ & $1984.984\pm244.580$ & $1.472\pm0.199$\\ \midrule
\multirow{2}{*}{B} & - & $0.861$ & $\mathbf{0.826}\pm0.032$ & $\mathbf{0.965}\pm0.008$ & $\mathbf{176.483}\pm33.399$ & $\mathbf{0.844}\pm0.154$\\
~ & LS & $\mathbf{0.869}$ & $0.320\pm0.062$ & $0.602\pm0.068$ & $233.413\pm43.395$ & $1.107\pm0.186$ \\
\bottomrule
    \end{tabular}
    }
    \end{threeparttable}
\end{table}

\subsection{Validation for Defense Methods on Models with Low Accuracy}
\label{appendix: low-acc}

The results for models with low accuracy are listed in Table \ref{table:defense_lowacc}. Obviously, the defense is valid when the target model has relatively low prediction accuracy. 

\begin{table}[!ht]
    \setlength{\tabcolsep}{5pt}
    \normalsize
    \centering
    \caption{Evaluation on defense methods for models with low accuracy.}
    \label{table:defense_lowacc}
    \begin{threeparttable} 
    \resizebox{\linewidth}{!}{
    \begin{tabular}{cccccccc}
        \toprule
\textbf{Method} & \textbf{Hyperparameters} & \textbf{Test Acc} & $\uparrow{\mathbf{Acc}@1}$ & $\uparrow{\mathbf{Acc}@5}$ & $\downarrow\mathbf{\delta}_{eval}$ & $\downarrow\mathbf{\delta}_{face}$ & $\downarrow\textbf{FID}$\\ \midrule
NO Defense & - & 92.170 & 0.686 & 0.914 & 262.471 & 0.767 & 68.454 \\ 
MID & $\alpha=0.005$ & 88.240 & 0.568 & 0.820 & 246.021 & 0.757 & 69.663 \\
BiDO & $\alpha=0.01,\beta= 0.1$ & 88.620 & 0.582 & 0.874 & 275.453 & 0.793 & 68.248 \\
TL & $\alpha=0.4$ & 89.160 & 0.316 & 0.616 & 279.439 & 0.897 & 63.241 \\
        \bottomrule
    \end{tabular}
    }
    \end{threeparttable}
\end{table}

\begin{figure}[!ht]
\centerline{\includegraphics[width=.49\textwidth]{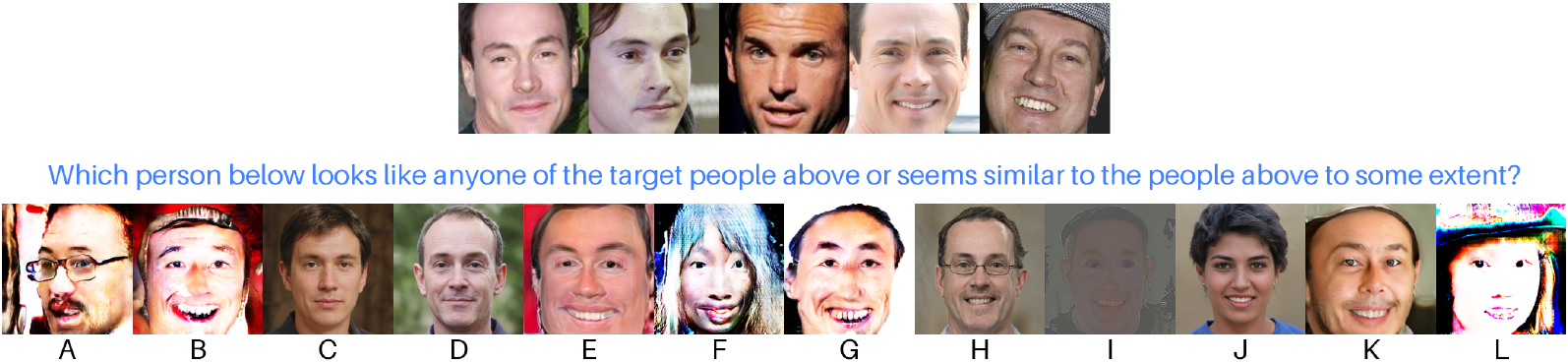}}
\caption{User interface of the user study.}
\label{usersample}
\end{figure}

\subsection{The abnormal visualization of Mirror}
\label{appendix: mirror-visual}

\begin{figure}[tbp]
\centerline{\includegraphics[width=\columnwidth]{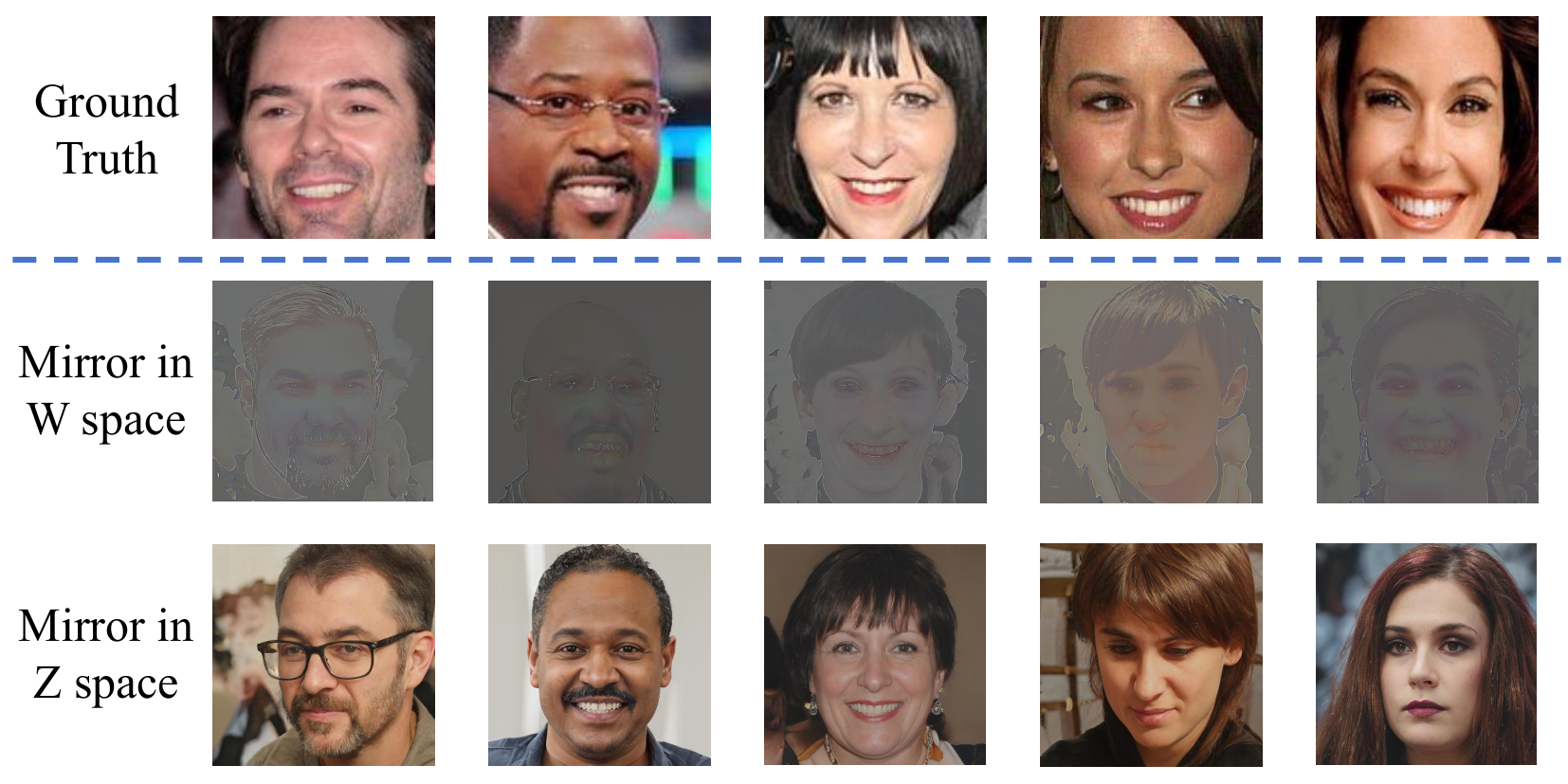}}
\caption{Visualization of Mirror in different search space.}
\label{mirror}
\end{figure}

To determine the cause of visualization anomalies in the Mirror method, we conduct a simple ablation experiment on its optimization objective, as denoted in Table \ref{table:mirror search space} and Fig \ref{mirror}. The experimental results demonstrate that when substituting the optimization objective with the $\mathcal{Z}$ space latent vectors, the visualization of Mirror returns to normality. Therefore, the underlying reason of the visualization anomalies is likely attributed to genetic-algorithm-induced corruption of StyleGAN's $\mathcal{W}$ latent vectors. As a disentangled latent vector that deeply participates in StyleGAN's hierarchical progressive image generation process, the compromised $\mathcal{W}$ latent vector ultimately leads to image artifacts in the generated outputs.
\begin{table}[!ht]
    \setlength{\tabcolsep}{5pt}
    \normalsize
    \centering
    \caption{Comparison of Mirror attack between different search space.}
    \label{table:mirror search space}
    \begin{threeparttable} 
    \resizebox{\linewidth}{!}{
    \begin{tabular}{ccccccc}
        \toprule
\textbf{Scenario} & \textbf{Space} & $\uparrow{\mathbf{Acc}@1}$ & $\uparrow{\mathbf{Acc}@5}$ & $\downarrow\mathbf{\delta}_{eval}$ & $\downarrow\mathbf{\delta}_{face}$ \\ \midrule
\multirow{2}{*}{White-box}& $\mathcal{W}$ & $0.348\pm0.023$ & $0.649\pm0.016$ & $197.741\pm32.212$ & $1.049\pm0.154$\\
& $\mathcal{Z}$ & $0.206\pm 0.013$ & $0.462\pm 0.020$ & $276.824\pm 551.751$ & $1.058\pm 0.147$ \\ \midrule
\multirow{2}{*}{Black-box}& $\mathcal{W}$ & $0.611\pm0.051$ & $0.862\pm0.018$ & $198.609\pm40.255$ & $1.049\pm0.192$ \\
& $\mathcal{Z}$ & $0.628\pm 0.025$ & $0.880\pm 0.011$ & $177.098\pm 34.300$ & $0.965\pm 0.191$ \\

\bottomrule
    \end{tabular}
    }
    \end{threeparttable}
\end{table}

\subsection{User Study}
\label{appendix: user-study}

In this section, we conduct a user study considering a subjective evaluation of MI attacks to further evaluate their effectiveness. We use different attack results in the $224\times224$ resolution scenarios where the public dataset is FFHQ and the private dataset is FaceScrub. We randomly select $60$ classes for user study and publish survey questionnaires on the \href{www.credamo.world}{credamo} website. The user interface is shown in Fig. \ref{usersample}. For each class, users are shown $5$ private images (above in Fig.\ref{usersample}) as examples and select the reconstructed images (below in Fig.\ref{usersample}) that have the top-3 visual similarity to the examples. In total, $160$ users have participated in the annotation, and the result is shown in Table \ref{tab:user study result}. 

\begin{table}[!ht]
    \setlength{\tabcolsep}{5pt}
    \normalsize
    \centering
    \caption{The results of user study under different attacks.}
    \label{tab:user study result}
    \begin{threeparttable} 
    \resizebox{\linewidth}{!}{
    \begin{tabular}{cccc}
        \toprule
\textbf{Attacks} & $\uparrow\textbf{Proportion of samples selected}$   \\ \midrule
\textbf{GMI} & 0.035 \\
\textbf{KEDMI} & 0.084 \\
\textbf{Mirror(white-box)} & 0.493 \\
\textbf{PPA} & 0.564 \\
\textbf{PLGMI} & 0.326 \\
\textbf{LOMMA+GMI} & 0.074 \\
\textbf{LOMMA+KEDMA} & 0.083 \\
\textbf{IF-GMI} & 0.419 \\ \midrule
\textbf{BREPMI} & 0.026 \\
\textbf{Mirror(black-box)} & 0.027 \\
\textbf{C2FMI} & 0.066 \\
\textbf{LOKT} & 0.059 \\
\bottomrule
    \end{tabular}
}
    \end{threeparttable}
\end{table}

\label{limitation}
\section{Limitations and Future Plans.}
During the writing process, there were new works about MI published in the top-tier conferences, such as \cite{ppdg}. As a toolbox for MI research, our benchmark is continuously updated and all the new works are scheduled to be implemented in the near future. In addition to developing new algorithms, it is also essential to conduct further research on the MI attacks and defenses to make in-depth analysis about their characteristics and bring valuable new insights.

\begin{thebibliography}{65}
\providecommand{\natexlab}[1]{#1}
\providecommand{\url}[1]{\texttt{#1}}
\expandafter\ifx\csname urlstyle\endcsname\relax
  \providecommand{\doi}[1]{doi: #1}\else
  \providecommand{\doi}{doi: \begingroup \urlstyle{rm}\Url}\fi

\bibitem[An et~al.(2022)An, Tao, Xu, Liu, Shen, Yao, Xu, and Zhang]{mirror}
Shengwei An, Guanhong Tao, Qiuling Xu, Yingqi Liu, Guangyu Shen, Yuan Yao, Jingwei Xu, and Xiangyu Zhang.
\newblock Mirror: Model inversion for deep learning network with high fidelity.
\newblock In \emph{Proceedings of the Network and Distributed System Security Symposium}, 2022.

\bibitem[Chen et~al.(2021{\natexlab{a}})Chen, Kahla, Jia, and Qi]{chen2021knowledge}
Si Chen, Mostafa Kahla, Ruoxi Jia, and Guo-Jun Qi.
\newblock Knowledge-enriched distributional model inversion attacks.
\newblock In \emph{Proceedings of the IEEE/CVF international conference on computer vision}, pages 16178--16187, 2021{\natexlab{a}}.

\bibitem[Chen et~al.(2021{\natexlab{b}})Chen, Kahla, Jia, and Qi]{ked}
Si Chen, Mostafa Kahla, Ruoxi Jia, and Guo-Jun Qi.
\newblock Knowledge-enriched distributional model inversion attacks.
\newblock In \emph{Proceedings of the IEEE/CVF international conference on computer vision}, 2021{\natexlab{b}}.

\bibitem[Choi et~al.(2020)Choi, Uh, Yoo, and Ha]{afhq}
Yunjey Choi, Youngjung Uh, Jaejun Yoo, and Jung-Woo Ha.
\newblock Stargan v2: Diverse image synthesis for multiple domains.
\newblock In \emph{Proceedings of the IEEE/CVF conference on computer vision and pattern recognition}, pages 8188--8197, 2020.

\bibitem[Dataset(2011{\natexlab{a}})]{standforddog}
E Dataset.
\newblock Novel datasets for fine-grained image categorization.
\newblock In \emph{First Workshop on Fine Grained Visual Categorization, CVPR. Citeseer. Citeseer}. Citeseer, 2011{\natexlab{a}}.

\bibitem[Dataset(2011{\natexlab{b}})]{stanforddog}
E Dataset.
\newblock Novel datasets for fine-grained image categorization.
\newblock In \emph{First workshop on fine grained visual categorization, CVPR. Citeseer. Citeseer. Citeseer}, page~2. Citeseer, 2011{\natexlab{b}}.

\bibitem[Deng et~al.(2009)Deng, Dong, Socher, Li, Li, and Fei-Fei]{imagenet}
Jia Deng, Wei Dong, Richard Socher, Li-Jia Li, Kai Li, and Li Fei-Fei.
\newblock Imagenet: A large-scale hierarchical image database.
\newblock In \emph{2009 IEEE conference on computer vision and pattern recognition}, pages 248--255. Ieee, 2009.

\bibitem[Fang et~al.(2023)Fang, Chen, Wang, Wang, and Xia]{fang2023gifd}
Hao Fang, Bin Chen, Xuan Wang, Zhi Wang, and Shu-Tao Xia.
\newblock Gifd: A generative gradient inversion method with feature domain optimization.
\newblock In \emph{Proceedings of the IEEE/CVF International Conference on Computer Vision}, pages 4967--4976, 2023.

\bibitem[Fang et~al.(2024)Fang, Qiu, Yu, Yu, Kong, Chong, Chen, Wang, and Xia]{fang2024privacy}
Hao Fang, Yixiang Qiu, Hongyao Yu, Wenbo Yu, Jiawei Kong, Baoli Chong, Bin Chen, Xuan Wang, and Shu-Tao Xia.
\newblock Privacy leakage on dnns: A survey of model inversion attacks and defenses.
\newblock \emph{arXiv preprint arXiv:2402.04013}, 2024.

\bibitem[Feng et~al.(2022)Feng, Zhao, Fang, Feng, Wei, Li, and Shao]{feng2022h}
Shanshan Feng, Kaiqi Zhao, Lanting Fang, Kaiyu Feng, Wei Wei, Xutao Li, and Ling Shao.
\newblock H-diffu: hyperbolic representations for information diffusion prediction.
\newblock \emph{IEEE Transactions on Knowledge and Data Engineering}, 2022.

\bibitem[Fredrikson et~al.(2015)Fredrikson, Jha, and Ristenpart]{fredrikson2015model}
Matt Fredrikson, Somesh Jha, and Thomas Ristenpart.
\newblock Model inversion attacks that exploit confidence information and basic countermeasures.
\newblock In \emph{Proceedings of the ACM SIGSAC Conference on Computer and Communications Security}, pages 1322--1333, 2015.

\bibitem[Gong et~al.(2023)Gong, Wang, Li, Chen, and Wang]{gong2023gan}
Xueluan Gong, Ziyao Wang, Shuaike Li, Yanjiao Chen, and Qian Wang.
\newblock A gan-based defense framework against model inversion attacks.
\newblock \emph{IEEE Transactions on Information Forensics and Security}, 2023.

\bibitem[Goodfellow et~al.(2014{\natexlab{a}})Goodfellow, Pouget-Abadie, Mirza, Xu, Warde-Farley, Ozair, Courville, and Bengio]{goodfellow2014generative}
Ian Goodfellow, Jean Pouget-Abadie, Mehdi Mirza, Bing Xu, David Warde-Farley, Sherjil Ozair, Aaron Courville, and Yoshua Bengio.
\newblock Generative adversarial nets.
\newblock \emph{Advances in neural information processing systems}, 27, 2014{\natexlab{a}}.

\bibitem[Goodfellow et~al.(2014{\natexlab{b}})Goodfellow, Shlens, and Szegedy]{fgsm}
Ian~J Goodfellow, Jonathon Shlens, and Christian Szegedy.
\newblock Explaining and harnessing adversarial examples.
\newblock \emph{arXiv preprint arXiv:1412.6572}, 2014{\natexlab{b}}.

\bibitem[Gu et~al.(2020)Gu, Shen, and Zhou]{gu2020image}
Jinjin Gu, Yujun Shen, and Bolei Zhou.
\newblock Image processing using multi-code gan prior.
\newblock In \emph{Proceedings of the IEEE/CVF conference on computer vision and pattern recognition}, pages 3012--3021, 2020.

\bibitem[Han et~al.(2023)Han, Choi, Lee, and Kim]{rlb}
Gyojin Han, Jaehyun Choi, Haeil Lee, and Junmo Kim.
\newblock Reinforcement learning-based black-box model inversion attacks.
\newblock In \emph{Proceedings of the IEEE/CVF Conference on Computer Vision and Pattern Recognition}, 2023.

\bibitem[He et~al.(2016)He, Zhang, Ren, and Sun]{resnet}
Kaiming He, Xiangyu Zhang, Shaoqing Ren, and Jian Sun.
\newblock Deep residual learning for image recognition.
\newblock In \emph{Proceedings of the IEEE conference on computer vision and pattern recognition}, pages 770--778, 2016.

\bibitem[He(2020)]{hdcelebacropper}
Zhenliang He.
\newblock Hd celeba cropper.
\newblock \url{https://github.com/LynnHo/HD-CelebA-Cropper}, 2020.

\bibitem[Heusel et~al.(2017)Heusel, Ramsauer, Unterthiner, Nessler, and Hochreiter]{fid}
Martin Heusel, Hubert Ramsauer, Thomas Unterthiner, Bernhard Nessler, and Sepp Hochreiter.
\newblock Gans trained by a two time-scale update rule converge to a local nash equilibrium.
\newblock \emph{Advances in neural information processing systems}, 30, 2017.

\bibitem[Ho et~al.(2024)Ho, Hao, Chandrasegaran, Nguyen, and Cheung]{tl}
Sy-Tuyen Ho, Koh~Jun Hao, Keshigeyan Chandrasegaran, Ngoc-Bao Nguyen, and Ngai-Man Cheung.
\newblock Model inversion robustness: Can transfer learning help?
\newblock \emph{arXiv preprint arXiv:2405.05588}, 2024.

\bibitem[Kahla et~al.(2022)Kahla, Chen, Just, and Jia]{brep}
Mostafa Kahla, Si Chen, Hoang~Anh Just, and Ruoxi Jia.
\newblock Label-only model inversion attacks via boundary repulsion.
\newblock In \emph{Proceedings of the IEEE/CVF Conference on Computer Vision and Pattern Recognition}, 2022.

\bibitem[Karras et~al.(2019)Karras, Laine, and Aila]{karras2019style}
Tero Karras, Samuli Laine, and Timo Aila.
\newblock A style-based generator architecture for generative adversarial networks.
\newblock In \emph{Proceedings of the IEEE/CVF conference on computer vision and pattern recognition}, pages 4401--4410, 2019.

\bibitem[Karras et~al.(2020{\natexlab{a}})Karras, Aittala, Hellsten, Laine, Lehtinen, and Aila]{metfaces}
Tero Karras, Miika Aittala, Janne Hellsten, Samuli Laine, Jaakko Lehtinen, and Timo Aila.
\newblock Training generative adversarial networks with limited data.
\newblock \emph{Advances in neural information processing systems}, 33:\penalty0 12104--12114, 2020{\natexlab{a}}.

\bibitem[Karras et~al.(2020{\natexlab{b}})Karras, Laine, Aittala, Hellsten, Lehtinen, and Aila]{stylegan2}
Tero Karras, Samuli Laine, Miika Aittala, Janne Hellsten, Jaakko Lehtinen, and Timo Aila.
\newblock Analyzing and improving the image quality of stylegan.
\newblock In \emph{Proceedings of the IEEE/CVF conference on computer vision and pattern recognition}, pages 8110--8119, 2020{\natexlab{b}}.

\bibitem[Koh et~al.(2024)Koh, Ho, Nguyen, and Cheung]{TTS}
Jun~Hao Koh, Sy-Tuyen Ho, Ngoc-Bao Nguyen, and Ngai-man Cheung.
\newblock On the vulnerability of skip connections to model inversion attacks.
\newblock \emph{arXiv preprint arXiv:2409.01696}, 2024.

\bibitem[Kurakin et~al.(2018)Kurakin, Goodfellow, and Bengio]{bim}
Alexey Kurakin, Ian~J Goodfellow, and Samy Bengio.
\newblock Adversarial examples in the physical world.
\newblock In \emph{Artificial intelligence safety and security}, pages 99--112. Chapman and Hall/CRC, 2018.

\bibitem[Kynk{\"a}{\"a}nniemi et~al.(2019)Kynk{\"a}{\"a}nniemi, Karras, Laine, Lehtinen, and Aila]{pr}
Tuomas Kynk{\"a}{\"a}nniemi, Tero Karras, Samuli Laine, Jaakko Lehtinen, and Timo Aila.
\newblock Improved precision and recall metric for assessing generative models.
\newblock \emph{Advances in Neural Information Processing Systems}, 32, 2019.

\bibitem[LeCun et~al.(1989)LeCun, Boser, Denker, Henderson, Howard, Hubbard, and Jackel]{lecun1989handwritten}
Yann LeCun, Bernhard Boser, John Denker, Donnie Henderson, Richard Howard, Wayne Hubbard, and Lawrence Jackel.
\newblock Handwritten digit recognition with a back-propagation network.
\newblock \emph{Advances in neural information processing systems}, 2, 1989.

\bibitem[Li et~al.(2022)Li, Rakin, Chen, He, Fan, and Chakrabarti]{li2022ressfl}
Jingtao Li, Adnan~Siraj Rakin, Xing Chen, Zhezhi He, Deliang Fan, and Chaitali Chakrabarti.
\newblock Ressfl: A resistance transfer framework for defending model inversion attack in split federated learning.
\newblock In \emph{Proceedings of the IEEE/CVF Conference on Computer Vision and Pattern Recognition}, pages 10194--10202, 2022.

\bibitem[Liu et~al.(2015)Liu, Luo, Wang, and Tang]{celeba}
Ziwei Liu, Ping Luo, Xiaogang Wang, and Xiaoou Tang.
\newblock Deep learning face attributes in the wild.
\newblock In \emph{Proceedings of the IEEE international conference on computer vision}, pages 3730--3738, 2015.

\bibitem[M{\k{a}}dry et~al.(2017)M{\k{a}}dry, Makelov, Schmidt, Tsipras, and Vladu]{pgd}
Aleksander M{\k{a}}dry, Aleksandar Makelov, Ludwig Schmidt, Dimitris Tsipras, and Adrian Vladu.
\newblock Towards deep learning models resistant to adversarial attacks.
\newblock \emph{stat}, 1050\penalty0 (9), 2017.

\bibitem[Naeem et~al.(2020)Naeem, Oh, Uh, Choi, and Yoo]{dc}
Muhammad~Ferjad Naeem, Seong~Joon Oh, Youngjung Uh, Yunjey Choi, and Jaejun Yoo.
\newblock Reliable fidelity and diversity metrics for generative models.
\newblock In \emph{International Conference on Machine Learning}, pages 7176--7185. PMLR, 2020.

\bibitem[Ng and Winkler(2014)]{facescrub}
Hong-Wei Ng and Stefan Winkler.
\newblock A data-driven approach to cleaning large face datasets.
\newblock In \emph{2014 IEEE international conference on image processing (ICIP)}, pages 343--347, 2014.

\bibitem[Nguyen et~al.(2023{\natexlab{a}})Nguyen, Chandrasegaran, Abdollahzadeh, and Cheung]{lokt}
Ngoc-Bao Nguyen, Keshigeyan Chandrasegaran, Milad Abdollahzadeh, and Ngai-Man Cheung.
\newblock Label-only model inversion attacks via knowledge transfer.
\newblock \emph{arXiv preprint arXiv:2310.19342}, 2023{\natexlab{a}}.

\bibitem[Nguyen et~al.(2023{\natexlab{b}})Nguyen, Chandrasegaran, Abdollahzadeh, and Cheung]{lomma}
Ngoc-Bao Nguyen, Keshigeyan Chandrasegaran, Milad Abdollahzadeh, and Ngai-Man Cheung.
\newblock Re-thinking model inversion attacks against deep neural networks.
\newblock In \emph{Proceedings of the IEEE/CVF Conference on Computer Vision and Pattern Recognition}, pages 16384--16393, 2023{\natexlab{b}}.

\bibitem[Ozbayoglu et~al.(2020)Ozbayoglu, Gudelek, and Sezer]{ozbayoglu2020deep}
Ahmet~Murat Ozbayoglu, Mehmet~Ugur Gudelek, and Omer~Berat Sezer.
\newblock Deep learning for financial applications: A survey.
\newblock \emph{Applied soft computing}, 93:\penalty0 106384, 2020.

\bibitem[Peng et~al.(2022)Peng, Liu, Zhang, Lan, Ye, Liu, and Han]{bido}
Xiong Peng, Feng Liu, Jingfeng Zhang, Long Lan, Junjie Ye, Tongliang Liu, and Bo Han.
\newblock Bilateral dependency optimization: Defending against model-inversion attacks.
\newblock In \emph{Proceedings of the 28th ACM SIGKDD Conference on Knowledge Discovery and Data Mining}, pages 1358--1367, 2022.

\bibitem[Peng et~al.(2025)Peng, Han, Liu, Liu, and Zhou]{ppdg}
Xiong Peng, Bo Han, Feng Liu, Tongliang Liu, and Mingyuan Zhou.
\newblock Pseudo-private data guided model inversion attacks.
\newblock \emph{Advances in Neural Information Processing Systems}, 37:\penalty0 33338--33375, 2025.

\bibitem[Qiu et~al.(2024)Qiu, Fang, Yu, Chen, Qiu, and Xia]{ifgmi}
Yixiang Qiu, Hao Fang, Hongyao Yu, Bin Chen, MeiKang Qiu, and Shu-Tao Xia.
\newblock A closer look at gan priors: Exploiting intermediate features for enhanced model inversion attacks.
\newblock In \emph{European Conference on Computer Vision}, 2024.

\bibitem[Schroff et~al.(2015)Schroff, Kalenichenko, and Philbin]{schroff2015facenet}
Florian Schroff, Dmitry Kalenichenko, and James Philbin.
\newblock Facenet: A unified embedding for face recognition and clustering.
\newblock In \emph{Proceedings of the IEEE conference on computer vision and pattern recognition}, pages 815--823, 2015.

\bibitem[Simonyan and Zisserman(2014)]{vgg}
Karen Simonyan and Andrew Zisserman.
\newblock Very deep convolutional networks for large-scale image recognition.
\newblock \emph{arXiv preprint arXiv:1409.1556}, 2014.

\bibitem[Song et~al.(2017)Song, Ristenpart, and Shmatikov]{song2017machine}
Congzheng Song, Thomas Ristenpart, and Vitaly Shmatikov.
\newblock Machine learning models that remember too much.
\newblock In \emph{Proceedings of the ACM SIGSAC Conference on Computer and Communications Security}, pages 587--601, 2017.

\bibitem[Struppek et~al.(2022)Struppek, Hintersdorf, Correira, Adler, and Kersting]{ppa}
Lukas Struppek, Dominik Hintersdorf, Antonio De~Almeida Correira, Antonia Adler, and Kristian Kersting.
\newblock Plug \& play attacks: Towards robust and flexible model inversion attacks.
\newblock In \emph{International Conference on Machine Learning}, 2022.

\bibitem[Struppek et~al.(2024)Struppek, Hintersdorf, and Kersting]{ls}
Lukas Struppek, Dominik Hintersdorf, and Kristian Kersting.
\newblock Be careful what you smooth for: Label smoothing can be a privacy shield but also a catalyst for model inversion attacks.
\newblock In \emph{The Twelfth International Conference on Learning Representations}, 2024.

\bibitem[Su et~al.(2019)Su, Vargas, and Sakurai]{onepixel}
Jiawei Su, Danilo~Vasconcellos Vargas, and Kouichi Sakurai.
\newblock One pixel attack for fooling deep neural networks.
\newblock \emph{IEEE Transactions on Evolutionary Computation}, 23\penalty0 (5):\penalty0 828--841, 2019.

\bibitem[Szegedy et~al.(2016)Szegedy, Vanhoucke, Ioffe, Shlens, and Wojna]{inceptionv3}
Christian Szegedy, Vincent Vanhoucke, Sergey Ioffe, Jon Shlens, and Zbigniew Wojna.
\newblock Rethinking the inception architecture for computer vision.
\newblock In \emph{Proceedings of the IEEE conference on computer vision and pattern recognition}, pages 2818--2826, 2016.

\bibitem[Titcombe et~al.(2021)Titcombe, Hall, Papadopoulos, and Romanini]{titcombe2021practical}
Tom Titcombe, Adam~J Hall, Pavlos Papadopoulos, and Daniele Romanini.
\newblock Practical defences against model inversion attacks for split neural networks.
\newblock \emph{arXiv preprint arXiv:2104.05743}, 2021.

\bibitem[Tu et~al.(2022)Tu, Talebi, Zhang, Yang, Milanfar, Bovik, and Li]{tu2022maxvit}
Zhengzhong Tu, Hossein Talebi, Han Zhang, Feng Yang, Peyman Milanfar, Alan Bovik, and Yinxiao Li.
\newblock Maxvit: Multi-axis vision transformer.
\newblock In \emph{European conference on computer vision}, pages 459--479. Springer, 2022.

\bibitem[Wang et~al.(2021{\natexlab{a}})Wang, Fu, Li, Khisti, Zemel, and Makhzani]{vmi}
Kuan-Chieh Wang, Yan Fu, Ke Li, Ashish Khisti, Richard Zemel, and Alireza Makhzani.
\newblock Variational model inversion attacks.
\newblock In \emph{Advances in Neural Information Processing Systems}, 2021{\natexlab{a}}.

\bibitem[Wang et~al.(2021{\natexlab{b}})Wang, Fu, Li, Khisti, Zemel, and Makhzani]{wang2021variational}
Kuan-Chieh Wang, Yan Fu, Ke Li, Ashish Khisti, Richard Zemel, and Alireza Makhzani.
\newblock Variational model inversion attacks.
\newblock \emph{Advances in Neural Information Processing Systems}, 34:\penalty0 9706--9719, 2021{\natexlab{b}}.

\bibitem[Wang et~al.(2022)Wang, Lin, Yan, Chen, Ding, Song, and Meng]{wang2022wearable}
Peng Wang, Zihuai Lin, Xucun Yan, Zijiao Chen, Ming Ding, Yang Song, and Lu Meng.
\newblock A wearable ecg monitor for deep learning based real-time cardiovascular disease detection.
\newblock \emph{arXiv preprint arXiv:2201.10083}, 2022.

\bibitem[Wang et~al.(2021{\natexlab{c}})Wang, Zhang, and Jia]{mid}
Tianhao Wang, Yuheng Zhang, and Ruoxi Jia.
\newblock Improving robustness to model inversion attacks via mutual information regularization.
\newblock In \emph{Proceedings of the AAAI Conference on Artificial Intelligence}, pages 11666--11673, 2021{\natexlab{c}}.

\bibitem[Wen et~al.(2021)Wen, Yiu, and Hui]{wen2021defending}
Jing Wen, Siu-Ming Yiu, and Lucas~CK Hui.
\newblock Defending against model inversion attack by adversarial examples.
\newblock In \emph{2021 IEEE International Conference on Cyber Security and Resilience (CSR)}, pages 551--556. IEEE, 2021.

\bibitem[Xu et~al.(2022)Xu, Liu, Tegmark, and Jaakkola]{xu2022poisson}
Yilun Xu, Ziming Liu, Max Tegmark, and Tommi Jaakkola.
\newblock Poisson flow generative models.
\newblock \emph{Advances in Neural Information Processing Systems}, 35:\penalty0 16782--16795, 2022.

\bibitem[Yang et~al.(2019)Yang, Zhang, Chang, and Liang]{yang2019neural}
Ziqi Yang, Jiyi Zhang, Ee-Chien Chang, and Zhenkai Liang.
\newblock Neural network inversion in adversarial setting via background knowledge alignment.
\newblock In \emph{Proceedings of the ACM SIGSAC Conference on Computer and Communications Security}, 2019.

\bibitem[Yang et~al.(2020)Yang, Shao, Xuan, Chang, and Zhang]{yang2020defending}
Ziqi Yang, Bin Shao, Bohan Xuan, Ee-Chien Chang, and Fan Zhang.
\newblock Defending model inversion and membership inference attacks via prediction purification.
\newblock \emph{arXiv preprint arXiv:2005.03915}, 2020.

\bibitem[Yang et~al.(2023)Yang, Wang, Yang, Wan, Zhao, Chang, Zhang, and Ren]{yang2023purifier}
Ziqi Yang, Lijin Wang, Da Yang, Jie Wan, Ziming Zhao, Ee-Chien Chang, Fan Zhang, and Kui Ren.
\newblock Purifier: Defending data inference attacks via transforming confidence scores.
\newblock In \emph{Proceedings of the AAAI Conference on Artificial Intelligence}, pages 10871--10879, 2023.

\bibitem[Ye et~al.(2022)Ye, Shen, Zhu, Liu, and Zhou]{ye2022one}
Dayong Ye, Sheng Shen, Tianqing Zhu, Bo Liu, and Wanlei Zhou.
\newblock One parameter defense—defending against data inference attacks via differential privacy.
\newblock \emph{IEEE Transactions on Information Forensics and Security}, 17:\penalty0 1466--1480, 2022.

\bibitem[Ye et~al.(2023)Ye, Luo, Naseem, Yang, Shi, and Jia]{c2f}
Zipeng Ye, Wenjian Luo, Muhammad~Luqman Naseem, Xiangkai Yang, Yuhui Shi, and Yan Jia.
\newblock C2fmi: Corse-to-fine black-box model inversion attack.
\newblock \emph{IEEE Transactions on Dependable and Secure Computing}, 2023.

\bibitem[Yuan et~al.(2023{\natexlab{a}})Yuan, Chen, Zhang, Zhang, Yu, and Zhang]{plg}
Xiaojian Yuan, Kejiang Chen, Jie Zhang, Weiming Zhang, Nenghai Yu, and Yang Zhang.
\newblock Pseudo label-guided model inversion attack via conditional generative adversarial network.
\newblock In \emph{Proceedings of the AAAI Conference on Artificial Intelligence}, 2023{\natexlab{a}}.

\bibitem[Yuan et~al.(2023{\natexlab{b}})Yuan, Chen, Zhang, Zhang, Yu, and Zhang]{yuan2023pseudo}
Xiaojian Yuan, Kejiang Chen, Jie Zhang, Weiming Zhang, Nenghai Yu, and Yang Zhang.
\newblock Pseudo label-guided model inversion attack via conditional generative adversarial network.
\newblock In \emph{Proceedings of the AAAI Conference on Artificial Intelligence}, 2023{\natexlab{b}}.

\bibitem[Zhang et~al.(2022)Zhang, Wu, Zhang, Zhu, Lin, Zhang, Sun, He, Mueller, Manmatha, et~al.]{zhang2022resnest}
Hang Zhang, Chongruo Wu, Zhongyue Zhang, Yi Zhu, Haibin Lin, Zhi Zhang, Yue Sun, Tong He, Jonas Mueller, R Manmatha, et~al.
\newblock Resnest: Split-attention networks.
\newblock In \emph{Proceedings of the IEEE/CVF conference on computer vision and pattern recognition}, pages 2736--2746, 2022.

\bibitem[Zhang et~al.(2020{\natexlab{a}})Zhang, Jia, Pei, Wang, Li, and Song]{gmi}
Yuheng Zhang, Ruoxi Jia, Hengzhi Pei, Wenxiao Wang, Bo Li, and Dawn Song.
\newblock The secret revealer: Generative model-inversion attacks against deep neural networks.
\newblock In \emph{Proceedings of the IEEE/CVF Conference on Computer Vision and Pattern Recognition}, 2020{\natexlab{a}}.

\bibitem[Zhang et~al.(2020{\natexlab{b}})Zhang, Jia, Pei, Wang, Li, and Song]{zhang2020secret}
Yuheng Zhang, Ruoxi Jia, Hengzhi Pei, Wenxiao Wang, Bo Li, and Dawn Song.
\newblock The secret revealer: Generative model-inversion attacks against deep neural networks.
\newblock In \emph{Proceedings of the IEEE/CVF Conference on Computer Vision and Pattern Recognition}, 2020{\natexlab{b}}.

\bibitem[Zhuang et~al.(2025)Zhuang, Yu, Qiu, Fang, Chen, and Xia]{zhuangstealthy}
Tianqu Zhuang, Hongyao Yu, Yixiang Qiu, Hao Fang, Bin Chen, and Shu-Tao Xia.
\newblock Stealthy shield defense: A conditional mutual information-based approach against black-box model inversion attacks.
\newblock In \emph{International Conference on Learning Representations (ICLR)}, 2025.

\end{thebibliography}
\end{document}